%% file: main.tex
\documentclass[11pt]{article}

\usepackage[preprint]{acl}

\usepackage{times}
\usepackage{latexsym}

\usepackage[T1]{fontenc}

\usepackage[utf8]{inputenc}

\usepackage{microtype}

\usepackage{inconsolata}

\usepackage{graphicx}

\usepackage{booktabs}
\usepackage{multirow}
\usepackage{colortbl}
\usepackage{arydshln}
\usepackage{url}

\usepackage{amsmath}
\usepackage{amssymb}
\usepackage{bbm}
\usepackage{bm}

\usepackage{adjustbox}
\usepackage{subcaption}

\usepackage{xcolor}
\definecolor{backgray}{rgb}{0.9,0.9,0.9}
\definecolor{textgray}{rgb}{0.6,0.6,0.6}
\definecolor{diffgreen}{HTML}{00b06b}
\definecolor{diffred}{HTML}{ff4b00}
\definecolor{textred}{HTML}{e08080}

\usepackage{tcolorbox}
\usepackage{listings}

\title{Revisiting the Capacity Gap in Chain-of-Thought Distillation\\from a Practical Perspective}

\author{
  Tokio Kajitsuka$^{1}$\thanks{\,Equal contribution.}\thanks{\,This work was done during an internship at CyberAgent.}\qquad Ukyo Honda$^{2*}$\qquad Sho Takase$^{2}$\\
  $^1$\,The University of Tokyo, Tokyo, Japan\qquad $^2$\,CyberAgent, Tokyo, Japan\\
  \texttt{kajitsuka.tokio@gmail.com}\\
  \texttt{\{honda\_ukyo, takase\_sho\}@cyberagent.co.jp}
}

\begin{document}
\maketitle
\begin{abstract}
Chain-of-thought (CoT) distillation transfers reasoning behaviors from a strong teacher to a smaller student, but prior work reports a capacity gap: distillation may fail when the teacher--student capability mismatch is large.
We revisit the capacity gap from a practical perspective by re-examining commonly used experimental settings.
Notably, we find that CoT distillation often degrades performance compared to the student's pre-distillation baseline, and that some settings used in prior work, while suitable for establishing the capacity gap as a phenomenon, do not reflect realistic deployment scenarios.
Complementing prior work that establishes the capacity gap, we evaluate its practical impact under more realistic settings and find that it does not consistently dominate; stronger teachers tend to be preferable when candidate teachers differ substantially in performance.
Our results offer practical guidance for selecting teacher--student pairs in CoT distillation.
\end{abstract}

\section{Introduction}
\label{sec:intro}

Large language models (LLMs) demonstrate strong reasoning when prompted to generate intermediate steps \citep[\textbf{chain-of-thought (CoT)} prompting;][]{wei2022chain,kojima2022large}, but the resulting long outputs raise deployment costs.
\textbf{CoT distillation} mitigates this by training a smaller student to mimic the reasoning of a larger teacher \citep{ho-etal-2023-large,li-etal-2023-symbolic,pmlr-v202-fu23d,hsieh-etal-2023-distilling,magister-etal-2023-teaching,shridhar-etal-2023-distilling}.\footnote{
CoT distillation typically targets task-specific reasoning transfer, as opposed to large-scale pretraining-time distillation, which aims at broad capability acquisition.
}

A widely cited concern is the \textbf{capacity gap}: distillation may fail when the teacher--student capability mismatch is large \citep{li-etal-2025-small-models,chen-etal-2025-unveiling-key}, implying that the strongest teacher is not always optimal and that practitioners face a costly hyperparameter search over candidate teachers.

\looseness=-1
In this paper, we revisit the capacity gap from a practical perspective.
\textbf{Our aim is not to dispute the existence of the capacity gap, but to provide a complementary assessment of how much it matters in a realistic scenario}: given a task that the small, efficient student cannot adequately solve without additional training, which teacher should be selected to distill into the student?
We identify two protocol-level issues that obscure this practitioner's question in prior work: \textbf{(1) configurations are compared only \emph{after} distillation, without checking whether distillation improves over the pre-distillation baseline} \citep{li-etal-2025-small-models,chen-etal-2025-unveiling-key}, and we find that it often \emph{degrades} performance; and \textbf{(2) some experimental settings do not reflect realistic deployment scenarios}: \emph{cross-teacher data filtering} keeps only examples that all candidate teachers solve, stripping stronger teachers of correct data that only they provide, and \emph{larger-student settings} evaluate students larger than their teacher, deviating from the efficiency motivation of distillation \citep{li-etal-2025-small-models}.

By adding pre-distillation baselines and removing these constraints, we find that the capacity gap does not consistently dominate across tasks and settings, and that when candidate teachers differ substantially in performance, the benefits of the stronger teacher outweigh capacity-gap effects.

Our primary contribution is thus a practical re-examination and revision of evaluation protocols for CoT distillation, complementary to prior capacity gap studies, with the teacher-selection guideline as its conditional consequence.
Specifically, our contributions are as follows:
\begin{itemize}
    \item We identify pitfalls in existing evaluation protocols from a practical perspective, most notably the missing comparison against the pre-distillation baseline, which obscures cases where distillation \emph{degrades} performance.
    \item Through systematic experiments across model families under a more realistic protocol, we show that capacity-gap effects do not consistently dominate across tasks and settings.
    \item We derive conditional guidelines for practitioners: first verify that distillation improves over the pre-distillation baseline, and \emph{when} this is confirmed \emph{and} candidate teachers differ substantially in performance, prefer the higher-performing teacher.
\end{itemize}

\section{Preliminaries}
\label{sec:pre}

\subsection{CoT Distillation}
\label{sec:pre-cot-distill}

CoT distillation aims to transfer the reasoning capabilities of a large \emph{teacher} model to a smaller, more efficient \emph{student} model.
The general procedure consists of two stages.

\paragraph{Rationale Generation.}
\label{sec:pre-rationale}
Given an original training dataset $\mathcal{D} = \{(x_i, y_i)\}_{i=1}^{N}$ with ground-truth labels $y_i$, the teacher model $\mathcal{T}$ generates reasoning chains (rationales) $\{\hat{r}_i\}_{i=1}^{N}$ along with predicted answers $\{\hat{y}_i\}_{i=1}^{N}$. To ensure the quality of reasoning chains, only samples where the predicted answer matches the ground-truth label are retained:
\begin{equation}
    \mathcal{D}_{\text{CoT}} = \{(x_i, \hat{r}_i, \hat{y}_i) \mid (x_i, y_i) \in \mathcal{D}, \ \hat{y}_i = y_i\}.
\end{equation}

\paragraph{Student Fine-tuning.}
\label{sec:pre-ft}
The student model $\mathcal{S}$ is fine-tuned on the distillation dataset $\mathcal{D}_{\text{CoT}}$ to learn to produce similar reasoning chains. Formally, the student is trained to minimize the negative log-likelihood:
\begin{equation}
    \mathcal{L} = -\sum_{(x, \hat{r}, \hat{y}) \in \mathcal{D}_{\text{CoT}}} \log P_{\mathcal{S}}(\hat{r}, \hat{y} \mid x),
\end{equation}
where $P_{\mathcal{S}}$ denotes the probability distribution of the student model.

\looseness=-1
This hard distillation can be viewed as a special case of sequence-level knowledge distillation \citep{kim-rush-2016-sequence}, where the teacher distribution is approximated via sampling rather than used directly as a soft distribution \citep{hinton2015distilling}.
Hard distillation is the dominant paradigm in CoT distillation because teacher logits are often unavailable from proprietary models, and tokenizer mismatches across model families preclude logit-level transfer \citep{ho-etal-2023-large,li-etal-2023-symbolic,hsieh-etal-2023-distilling,magister-etal-2023-teaching,shridhar-etal-2023-distilling}.

\subsection{Capacity Gap}
\label{sec:pre-capacity-gap}
The \emph{capacity gap} refers to the empirical observation that distillation may yield suboptimal results when there is a substantial capability mismatch between the teacher and student models.
Such mismatch can arise from various sources, including differences in model size or in reasoning style, e.g., short vs. long chain-of-thought \citep{li-etal-2025-small-models}.

\looseness=-1
This phenomenon has been reported in conventional soft knowledge distillation \citep{lopez-paz2016unifying,cho2019efficacy,mirzadeh2020improved,jafari-etal-2021-annealing}.
CoT distillation lies on the same spectrum as soft knowledge distillation but at the hard-label end (Section~\ref{sec:pre-ft}).
Capacity gap analyses developed for soft knowledge distillation therefore do not straightforwardly transfer, requiring independent empirical investigation.
Prior work has conducted such investigations and reported the same phenomenon in CoT distillation \citep{li-etal-2025-small-models,chen-etal-2025-unveiling-key}, suggesting that using CoT from the strongest teacher may not always be optimal.

However, the practical impact of the capacity gap under realistic deployment conditions remains unexamined.
Our work is therefore complementary to these prior studies: rather than disputing the existence of the gap, we focus on the hard CoT distillation setting and examine whether commonly used evaluation protocols reflect realistic deployment scenarios, thereby assessing how much the capacity gap actually matters in practice.

\section{Pitfalls in Existing Evaluation Protocols}
\label{sec:pitfalls}

We critically examine experimental protocols commonly used in prior capacity gap studies \citep{li-etal-2025-small-models,chen-etal-2025-unveiling-key} and identify three pitfalls that may lead to conclusions that do not directly translate to practical deployment scenarios.

\begin{figure*}[t]
\centering
\begin{subfigure}[t]{0.5\textwidth}
    \vspace{0pt}
    \centering
    \includegraphics[width=\linewidth,keepaspectratio]{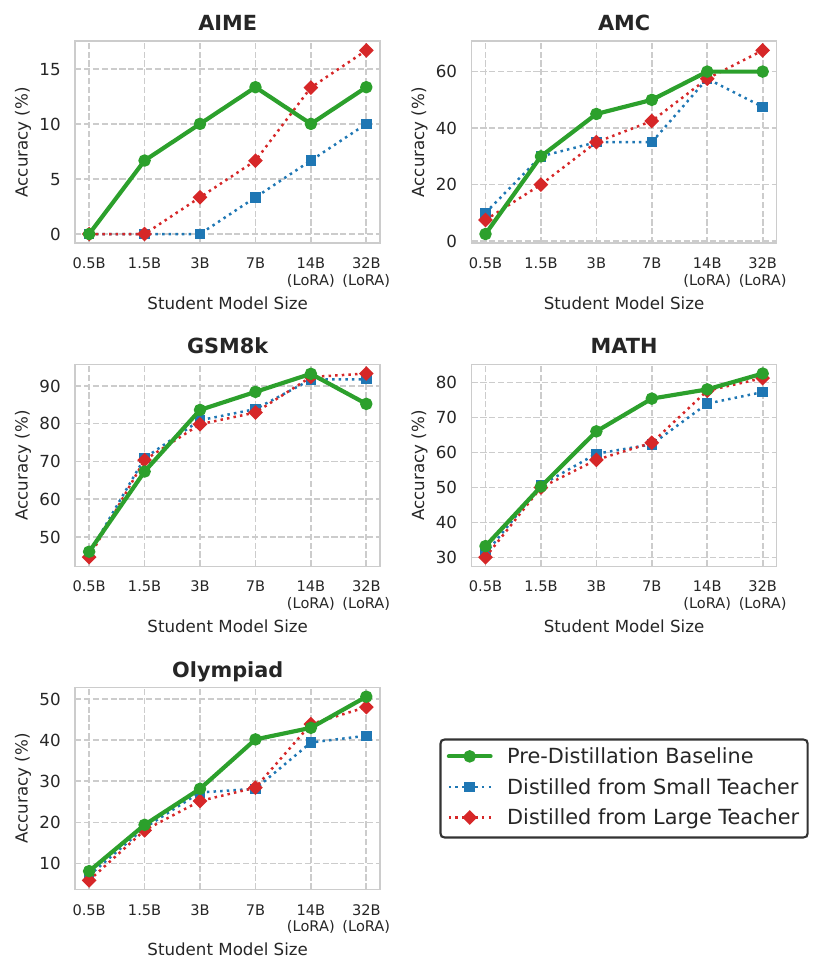}
    \caption{Results under the \emph{small--large} setting.}
    \label{fig:math-small-large}
\end{subfigure}%
\begin{subfigure}[t]{0.5\textwidth}
    \vspace{0pt}
    \centering
    \includegraphics[clip,trim=3.484375pt 0pt 3.484375pt 0pt,width=\linewidth,keepaspectratio]{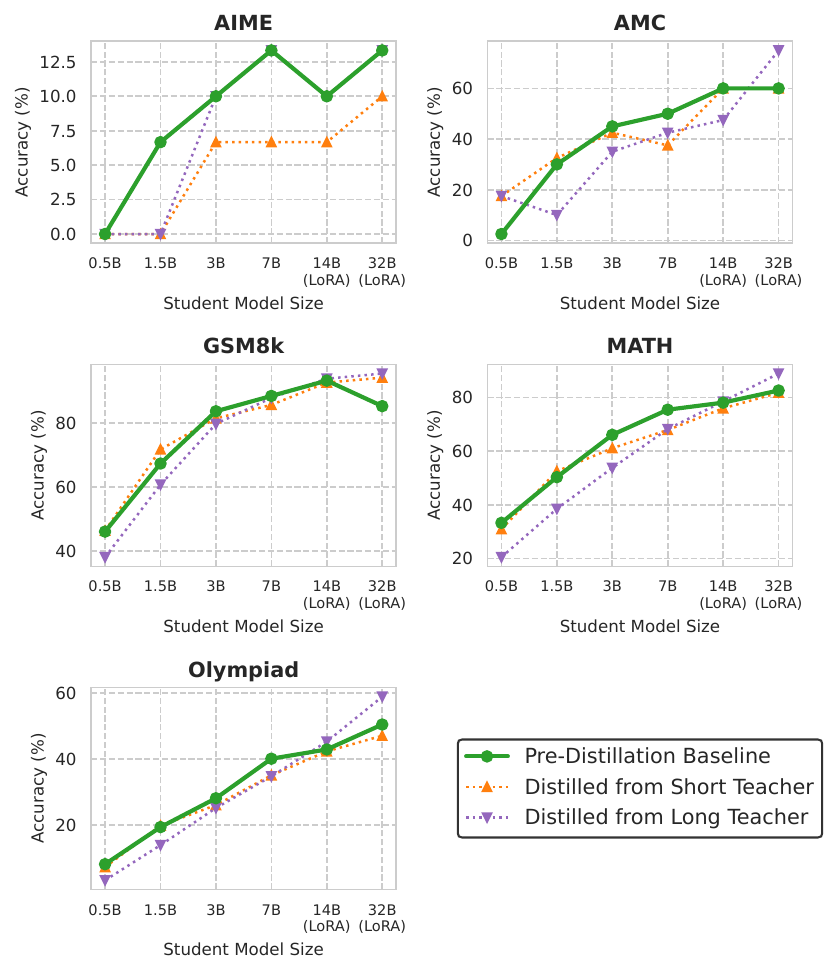}
    \caption{Results under the \emph{short--long} setting.}
    \label{fig:math-short-long}
\end{subfigure}
\caption{Performance comparison on five math benchmarks (AIME, AMC, GSM8K, MATH, and OlympiadBench).
We compare the pre-distillation baseline with distilled students.
(a) \emph{Small--large} setting: students distilled from Qwen2.5-3B-Instruct (Small Teacher) and Qwen2.5-72B-Instruct (Large Teacher).
(b) \emph{Short--long} setting: students distilled from Qwen2.5-32B-Instruct (Short Teacher) and QwQ-32B-Preview (Long Teacher).
}
\label{fig:math-results}
\end{figure*}

\subsection{Pitfall 1: Lack of Pre-Distillation Baseline Comparison}
\label{sec:pitfall-baseline}

Prior studies on the capacity gap \citep{li-etal-2025-small-models,chen-etal-2025-unveiling-key} compare different distillation configurations only \emph{after} distillation, without examining whether distillation itself improves upon the student model's pre-distillation performance.
This omission is problematic because recent LLMs already possess substantial reasoning capabilities and may already perform well on target tasks without fine-tuning.
Therefore, verifying that distillation actually \emph{improves} performance is essential for its practical adoption.

\paragraph{Experimental Setup.}
\label{sec:pitfall-baseline-setup}
To examine whether prior experiments constitute a valid scenario where distillation is expected to be beneficial, we add comparisons with pre-distillation baselines to the experimental settings of \citet{li-etal-2025-small-models}.
Our experiments are based on their publicly available code and data to ensure consistent experimental settings.\footnote{\url{https://github.com/Small-Model-Gap/Small-Model-Learnability-Gap}}
Below, we briefly summarize the experimental setup; see Appendix~\ref{sec:appendix-setup} for more details.

\emph{Training Data and Teacher Models.}
We use the distillation datasets publicly released by \citet{li-etal-2025-small-models}.
These datasets are constructed by generating CoT rationales for the MATH training set \citep{hendrycks2021measuring} and filtering to retain only examples where both teachers in each comparison produce correct answers (see Section~\ref{sec:pitfall-filter} for a discussion of the implications of this filtering).
Two teacher comparisons are considered: the \textbf{small--large} setting, which compares teachers of different model sizes, specifically CoT from Qwen2.5-3B-Instruct with Qwen2.5-72B-Instruct \citep{qwen2025qwen25}; and the \textbf{short--long} setting, which compares teachers that generate reasoning chains of different lengths, specifically short CoT from Qwen2.5-32B-Instruct with long CoT from QwQ-32B-Preview.

\emph{Evaluation.}
We evaluate the student models on five mathematical reasoning benchmarks of varying difficulty: MATH, GSM8K \citep{cobbe2021training}, AMC 2023, AIME 2024, and OlympiadBench \citep{he-etal-2024-olympiadbench}.
To ensure a fair comparison between the distilled students and the pre-distillation baseline, we apply the same evaluation prompt to all models.

\emph{Student Models and Training.}
We use Qwen2.5-\{0.5B, 1.5B, 3B, 7B, 14B, 32B\}-Instruct as student models and fine-tune them on the distillation datasets using \texttt{LLaMA-Factory} \citep{zheng-etal-2024-llamafactory}.
For all configurations, we use the same training hyperparameters as \citet{li-etal-2025-small-models}.

\paragraph{Results.}
\label{sec:pitfall-baseline-results}
Figures~\ref{fig:math-small-large} and \ref{fig:math-short-long} show the results in the small--large and short--long settings, respectively.
We found a striking result: in most configurations, CoT distillation \emph{degrades} performance compared to the student's pre-distillation baseline.
A plausible interpretation is that recent instruction-tuned LLMs already possess strong reasoning capabilities acquired through extensive pretraining, and fine-tuning on teacher-generated rationales does not necessarily improve, and can degrade, these existing capabilities.
We treat the specific mechanism as an open question; candidates include forgetting of pretrained knowledge, rationale-style mismatch, distribution shift, and teacher-CoT quality lower than that of the student's instruction-tuning data.\footnote{
As an initial probe into the mechanism, our analysis in Appendix~\ref{sec:appendix-artifacts} provides suggestive evidence: configurations with fewer parameter updates, and hence plausibly smaller drift from the initial parameters, suffer less degradation on MATH-500, which is consistent with an optimization-induced over-updating account, though it does not yet disentangle the candidate explanations above.
}
This indicates that the experimental setting does not represent a practical scenario where distillation is beneficial; comparing distillation strategies then amounts to comparing degrees of degradation rather than identifying effective approaches.\footnote{
We additionally found that such an impractical setting can lead to misleading artifacts: a strategy proposed in prior work to mitigate the capacity gap appeared effective, but this was largely due to hyperparameter discrepancies that reduced the number of parameter updates rather than achieving genuine improvements (see Appendix~\ref{sec:appendix-artifacts} for details).
}

\subsection{Pitfall 2: Data Filtering that Neutralizes Teacher Strength}
\label{sec:pitfall-filter}

\looseness=-1
When comparing different teacher models, \citet{li-etal-2025-small-models} filter the training data to include only examples where \emph{both} teachers produce correct answers.
This is a reasonable methodological choice for isolating the capacity gap: by controlling the training data, one can attribute performance differences purely to the quality of reasoning chains.
In this sense, prior work provides valid evidence that a capacity gap can exist under controlled conditions.
However, our goal is to assess \emph{how much the capacity gap matters in practice}.
A practitioner would use all examples that the teacher solves correctly, and stronger teachers then provide two advantages: more training examples due to higher accuracy, and coverage of harder cases that the other teacher fails to solve.
Cross-teacher filtering eliminates both.\footnote{
Importantly, cross-teacher filtering should not be confused with curating high-quality training data, which has proven effective for improving LLM reasoning \citep{ye2025limo}.
Quality-based selection retains examples based on their instructional value, whereas cross-teacher filtering merely restricts data to the intersection of what both teachers solve correctly, discarding correct examples without any quality-based rationale.
Indeed, our ablation study (Appendix~\ref{sec:appendix-filtering}) shows that \emph{applying cross-teacher filtering to our BBH experiments consistently degrades overall distillation performance}, confirming that this filtering reduces useful supervision rather than improving data quality.
}

\subsection{Pitfall 3: Inclusion of Larger-Student Settings}
\label{sec:pitfall-size}

Some prior experiments include configurations where the student model is \emph{larger} than the teacher model \citep{li-etal-2025-small-models}.
While such settings are valid for demonstrating the existence of the capacity gap, they do not align with the practical motivation for CoT distillation, whose primary goal is to reduce deployment costs by transferring knowledge to a smaller model.
Since our focus is on assessing the practical impact of the capacity gap, we restrict experiments to settings where the student is strictly smaller than the teacher.

\section{Re-Evaluation with Practical Protocols}
\label{sec:re-eval}

We now propose a revised evaluation protocol that addresses the pitfalls in Section~\ref{sec:pitfalls} and re-examine whether the capacity gap remains a concern under realistic conditions.

\subsection{Proposed Evaluation Protocol}
\label{sec:re-eval-protocol}

We propose a practical evaluation protocol consisting of the following modifications to existing experimental settings.

\paragraph{Modification 1: Task Selection Based on Expected Distillation Benefits.}
\label{sec:re-eval-protocol-task}
We first identify tasks where CoT distillation is likely to yield performance improvements, ensuring that we study scenarios where distillation is practically meaningful rather than comparing degrees of degradation.

\looseness=-1
We expect distillation to be most effective when the task requires knowledge beyond what pretraining provides.
Although many-shot in-context learning \citep[many-shot ICL;][]{agarwal2024manyshot} and distillation differ in mechanism, both aim to acquire task-specific knowledge from numerous examples; we therefore hypothesize that tasks where many-shot ICL is effective are also likely to benefit from distillation.
\citet{honda-etal-2025-distilling} report that many-shot ICL yields no gains on mathematical benchmarks with recent strong LLMs, but \textbf{BIG-Bench Hard} \citep[\textbf{BBH};][]{suzgun-etal-2023-challenging,srivastava2023beyond} remains an exception, likely because BBH tasks require task-specific patterns not inferable from prior knowledge alone.
Based on this reasoning, we select BBH as our benchmark, expecting that distillation will similarly benefit from the availability of numerous task-specific examples.

BBH consists of 23 diverse tasks, and we further narrow down to tasks where distillation has substantial room to improve the student.
Specifically, we measure the few-shot ICL performance of all teacher models and all pre-distillation student models on each task, and then compute the average performance within each group (i.e., the average across teacher models and the average across student models).
We then retain tasks where the difference between these averages exceeds a threshold.
This selection criterion ensures that there is sufficient room for the student to learn from the teacher, yielding tasks well-suited for examining CoT distillation in realistic scenarios.

\paragraph{Modification 2: Removal of Cross-Teacher Data Filtering.}
\label{sec:re-eval-protocol-filter}
Instead of filtering training data to the intersection of examples solved correctly by both weak and strong teachers (an analytical convention in prior work), we use each teacher's full set of correct examples, reflecting the natural choice a practitioner would make and letting stronger teachers benefit from their higher accuracy.
We acknowledge that this protocol does not fully isolate the capacity gap, since reasoning quality and data quantity vary together across teachers.
However, it suits our purpose of assessing the overall practical impact of teacher selection rather than its mechanistic breakdown.
As a complementary analysis, Appendix~\ref{sec:appendix-filtering} applies cross-teacher filtering and finds that the gains from stronger teachers are driven primarily by increased data quantity rather than by improved reasoning quality.

\paragraph{Modification 3: Restriction to Efficiency-Motivated Settings.}
\label{sec:re-eval-protocol-size}
We restrict our evaluation to configurations where the student model is smaller than the teacher model, aligning with the efficiency-driven motivation for CoT distillation.
This allows us to derive actionable insights for practical deployment.
The settings where knowledge is transferred from a smaller to a larger model, while theoretically interesting, fall outside the scope of this study.

\begin{figure*}[t]
\centering
\includegraphics[width=1.0\textwidth,keepaspectratio]{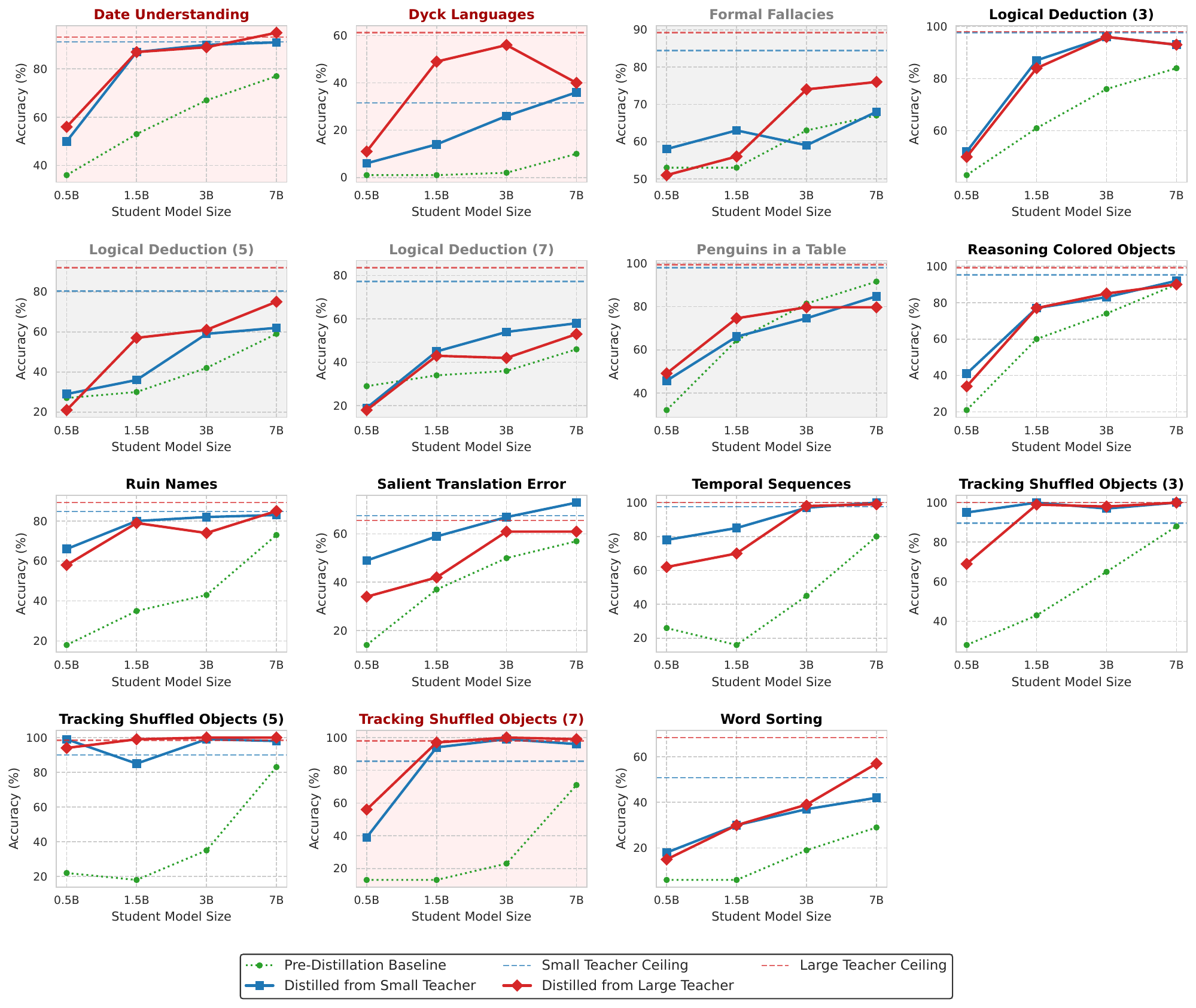}
\caption{
Results under the \emph{small--large} setting on 15 selected BBH tasks with our practical evaluation protocol.
We compare the pre-distillation baseline with students distilled from Qwen2.5-14B-Instruct (Small Teacher) and Qwen2.5-72B-Instruct (Large Teacher).
Teacher Ceiling indicates the few-shot performance of each teacher model.
Tasks with a \textcolor{textgray}{gray} background indicate cases where at least one distilled model underperformed the pre-distillation baseline.
Tasks with a \textcolor{textred}{red} background indicate cases where the results did not follow the capacity gap hypothesis.
}
\label{fig:bbh-small-large}
\end{figure*}

\begin{figure*}[t]
\centering
\includegraphics[width=1.0\textwidth,keepaspectratio]{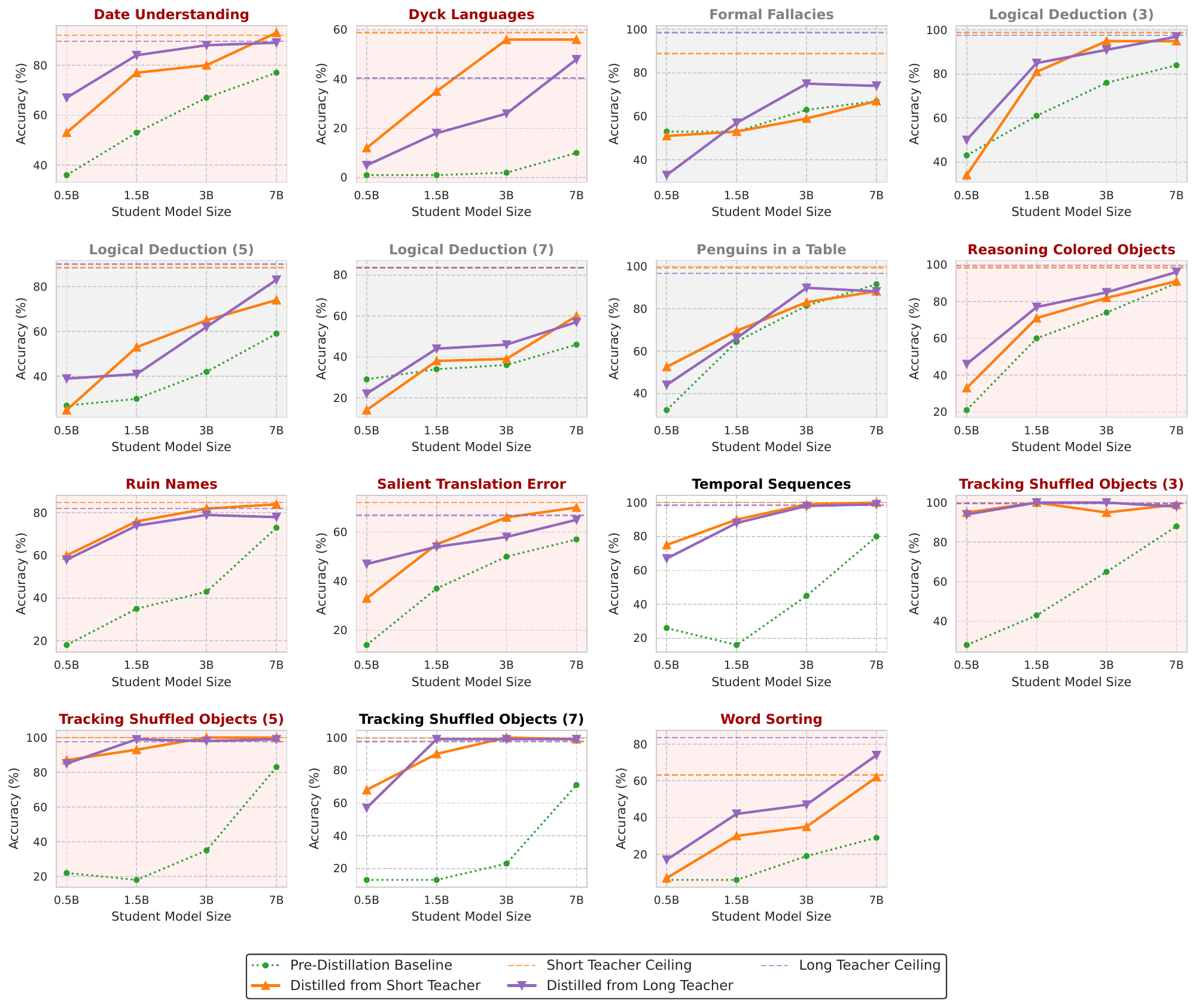}
\caption{
Results under the \emph{short--long} setting on 15 selected BBH tasks with our practical evaluation protocol.
We compare the pre-distillation baseline with students distilled from Qwen2.5-32B-Instruct (Short Teacher) and QwQ-32B-Preview (Long Teacher).
Teacher Ceiling indicates the few-shot performance of each teacher model.
Tasks with a \textcolor{textgray}{gray} background indicate cases where at least one distilled model underperformed the pre-distillation baseline.
Tasks with a \textcolor{textred}{red} background indicate cases where the results did not follow the capacity gap hypothesis.
}
\label{fig:bbh-short-long}
\end{figure*}

\subsection{Experimental Setup}
\label{sec:re-eval-setup}

Our experimental setup follows that of Section~\ref{sec:pitfall-baseline-setup}, with modifications according to the protocol described in Section~\ref{sec:re-eval-protocol}.
We select BBH tasks where the average few-shot accuracy gap between teacher and student model groups exceeds 30 percentage points, yielding 15 tasks (Appendix~\ref{sec:appendix-setup-bbh}).
For each task, we split examples into training and test sets in a 3:2 ratio.
To ensure sufficient student model sizes while keeping students smaller than teachers, we use Qwen2.5-14B-Instruct and Qwen2.5-72B-Instruct as small and large teachers in the \emph{small--large} setting; for \emph{short--long}, we use Qwen2.5-32B-Instruct and QwQ-32B-Preview as in Section~\ref{sec:pitfall-baseline-setup}.
We do not apply cross-teacher filtering; instead, for each teacher, we retain all examples where that teacher produces a correct answer.
For fair comparison, we apply the same few-shot CoT prompt to all models, including the pre-distillation baseline.
We also conduct cross-family experiments using Gemma-2 teachers with Qwen2.5 students to validate that our findings generalize beyond a single model family (Appendix~\ref{sec:appendix-gemma}).

\subsection{Results}
\label{sec:re-eval-results}
Figures~\ref{fig:bbh-small-large} and \ref{fig:bbh-short-long} present the results under our practical evaluation protocol.
We discuss the results further in Section~\ref{sec:guidelines}.
We additionally verify that these findings hold with Gemma-2 teachers in a cross-family setting (Appendix~\ref{sec:appendix-gemma}).

\paragraph{Distillation Efficacy.}
\label{sec:re-eval-results-efficacy}
In most configurations, distilled models outperform the pre-distillation baseline, confirming that many of the selected tasks represent valid distillation scenarios.
However, some tasks show limited or negative distillation effects: in the small--large setting, four tasks (Formal Fallacies, Logical Deduction (5) and (7), and Penguins in a Table), and in the short--long setting, five tasks (with the addition of Logical Deduction (3)).
We focus our subsequent analysis on the remaining tasks where distillation consistently improves performance (11 tasks in small--large, 10 tasks in short--long), and defer discussion of these problematic tasks to the end of this section.

\paragraph{Capacity Gap Effects.}
\label{sec:re-eval-results-gap}
Among tasks where distillation is effective, capacity gap effects are observed in some cases but their impact varies across tasks and settings.
We identify capacity gap effects using the following \emph{diagnostic pattern}: if the capacity gap were a dominant factor, less capable teachers would yield better students at the smallest student sizes, with this advantage diminishing or reversing as student size increases.\footnote{Because we report single-run results on relatively small per-task test sets (\hyperref[sec:limitations-statistical]{Limitations} section), per-task judgments should be interpreted with care.
However, this diagnostic is asymmetric in a way that protects our main claim: detecting a capacity gap requires the less capable teacher to \emph{actively} outperform the more capable one at small student sizes, so noise in marginal cases biases toward an inconclusive ``no clear gap'' reading rather than a false positive for capacity gap dominance.
A three-seed re-run on four representative tasks (Appendix~\ref{sec:appendix-seeds}) supports this reading: seed averaging leaves the task-level conclusions for these tasks unchanged, and no capacity gap pattern robust to seed variation emerges.
}
Using this criterion, in the small--large setting, we observe such effects in eight of the eleven distillation-effective tasks, while Date Understanding, Dyck Languages, and Tracking Shuffled Objects (7) show no such effects.
In the short--long setting, capacity gap effects appear only in Temporal Sequences and Tracking Shuffled Objects (7), while the remaining eight tasks show no such effects.

\paragraph{Effect of Teacher Performance Gap.}
\label{sec:re-eval-results-practical}
Although capacity gap effects are present in some configurations, we identify a consistent practical pattern: when there is a substantial performance gap \emph{between teacher models}, as observed in Dyck Languages and Word Sorting (Table~\ref{tab:bbh-pre-all}), higher-performing teachers produce better or comparable distilled students.
This finding suggests that, even when capacity gap effects exist, the benefits of stronger teachers outweigh these effects when the teacher performance difference is large.

\paragraph{Analysis of Ineffective Distillation Tasks.}
\label{sec:re-eval-results-ineffective}
For the tasks where distillation shows limited effectiveness (Formal Fallacies, Logical Deduction, Penguins in a Table), capacity gap patterns appear inconsistently across settings, and even for Formal Fallacies, where the teacher performance gap is relatively large, stronger teachers do not yield a consistent advantage.
Thus, when distillation itself is ineffective for a task, neither capacity gap considerations nor teacher strength provide reliable guidance for teacher selection.
However, since practitioners should not apply distillation to such tasks in the first place, this absence of guidance is not a practical concern.

\section{Discussion}
\label{sec:guidelines}
In this section, we synthesize our results into actionable recommendations for practitioners considering CoT distillation.

\paragraph{Guideline 1: Verify distillation efficacy against the pre-distillation baseline.}
\label{sec:guidelines-1}
Unlike earlier models that required task-specific training, recent LLMs possess substantial zero-shot and few-shot reasoning capabilities and may already perform well on target tasks without any fine-tuning.
Consequently, distillation is not guaranteed to improve performance and may even cause degradation.
Before deploying a distilled model or comparing distillation strategies, practitioners should confirm that distillation yields genuine improvements over the student's pre-distillation performance.

Since verifying this through actual fine-tuning is costly, a lightweight proxy is desirable.
Our results suggest that few-shot ICL performance gaps between teacher and student models may serve this purpose: in our BBH experiments, tasks selected based on large ICL gaps showed effective distillation in most configurations, indicating that this approach can potentially reduce the cost of identifying suitable tasks for CoT distillation.

Several pieces of evidence support the generality of this proxy beyond the setup used to derive it.
Across teacher families and data domains, applying the same 30 percentage-point threshold to our Gemma-2 cross-family experiments (Appendix~\ref{sec:appendix-gemma}) selects only tasks where distillation indeed improves over the baseline, and applying it to the math tasks in Section~\ref{sec:pitfall-baseline} yields no tasks above the threshold, consistent with our finding that distillation does not help there.
The threshold is also validated on both sides of the 30-point cutoff (Appendix~\ref{sec:appendix-threshold}): newly distilling three BBH tasks whose ICL gap falls just below the threshold (in the 20--30 point band) shows that distillation is not reliably effective on any of them, while most tasks selected under stricter cutoffs of 35 or 40 points remain distillation-effective (8 of 10 and 6 of 8, respectively, in both teacher settings).
Broader validation across more benchmarks and model families, as well as exploration of alternative selection criteria, remains an important direction for future work.

\paragraph{Guideline 2: If distillation is beneficial \emph{and} teacher performance differs substantially, prefer the higher-performing teacher.}
\label{sec:guidelines-2}
Our results indicate that when there is a substantial performance gap between candidate teachers, higher-performing teachers produce better or comparable distilled students, regardless of student model size.
We therefore state this as an explicitly conditional recommendation: \emph{if} distillation has been verified to improve over the pre-distillation baseline (Guideline~1) \emph{and} candidate teachers differ substantially in performance, \emph{then} practitioners should select the higher-performing teacher.
Where distillation itself is ineffective for a task, neither capacity gap considerations nor teacher strength provide reliable guidance for teacher selection, and the present guideline does not apply (Section~\ref{sec:re-eval-results-ineffective}).
Although this guideline may appear intuitive, prior capacity gap findings imply a costly search over weaker candidate teachers even when a substantially stronger one is available; showing that this search can be unnecessary in such cases is itself a practically useful outcome.

This practical benefit of stronger teachers has multiple components, of which increased training data, a natural consequence of higher teacher accuracy, is a major one.
Our ablation with cross-teacher filtering (Appendix~\ref{sec:appendix-filtering}) equalizes data sizes across teachers and shows that, in the two tasks where the stronger teacher provides substantially more data (Dyck Languages and Word Sorting, 1.3--1.8$\times$), the stronger-teacher advantage largely disappears, while most tasks with smaller data size differences (at most about $1.1\times$) remain largely unchanged.
A data size ratio of around $1.3\times$ may thus serve as a rough reference for when the stronger teacher's data-quantity advantage becomes practically decisive, though this is derived from a limited number of tasks and should be treated as an approximate heuristic rather than a rigid criterion.
When the performance gap is small, practitioners may consider other factors such as computational cost or reasoning chain length, as shorter chains reduce inference costs.\footnote{
Even when generation length is explicitly capped, capacity gap effects do not emerge consistently: in a budget-capped re-evaluation of the BBH students from our main experiments (Appendix~\ref{sec:appendix-budget}), students distilled from the short teacher prevail in some configurations under tight budgets, yet for the smallest (0.5B) student the long teacher's student stays ahead at every budget.
}

\section{Related Work}
\label{sec:related}

\subsection{Knowledge Distillation}
\label{sec:related-distill}
CoT distillation builds upon knowledge distillation \citep{hinton2015distilling} and sequence-level knowledge distillation \citep{kim-rush-2016-sequence}.
As discussed in Section~\ref{sec:pre-cot-distill}, hard distillation---supervised fine-tuning on teacher-generated sequences \citep{ho-etal-2023-large,li-etal-2023-symbolic,hsieh-etal-2023-distilling,magister-etal-2023-teaching,shridhar-etal-2023-distilling}---is the dominant paradigm.\footnote{While \citet{pmlr-v202-fu23d} partially employ soft distillation via KL divergence minimization, they fall back to hard distillation depending on tokenizer alignment.}
It has also been widely adopted for transferring general knowledge \citep{west-etal-2022-symbolic,mukherjee2023orca} or reasoning traces for test-time scaling \citep{muennighoff-etal-2025-s1}.

\subsection{Capacity Gap in Knowledge Distillation}
\label{sec:related-capacity-gap}
The capacity gap is also observed in conventional soft knowledge distillation \citep{lopez-paz2016unifying,cho2019efficacy,mirzadeh2020improved,jafari-etal-2021-annealing}.
These analyses, however, rely on access to the teacher's full output distribution (e.g., logit-based KL objectives or temperature annealing) and do not directly transfer to the hard CoT distillation studied here, where the student is trained only on sampled sequences (Section~\ref{sec:pre-ft}); the bridge is nonetheless tight, as \citet{kim-rush-2016-sequence} place sequence-level hard knowledge distillation on the same spectrum as soft knowledge distillation by approximating the teacher distribution via sampling, and the same phenomenon has been empirically reported in this regime \citep{li-etal-2025-small-models,chen-etal-2025-unveiling-key}.
\citet{zhang-etal-2025-towards-law} study the capacity gap in task-agnostic LM distillation, but their controlled setting, where the student is initialized by pruning the teacher, does not reflect practical scenarios involving independently pretrained models.
Our contribution is an empirical re-evaluation of the capacity gap in hard CoT distillation under practical conditions, \emph{complementary} to \citet{li-etal-2025-small-models} and \citet{chen-etal-2025-unveiling-key}: prior work establishes the gap under controlled protocols, whereas we assess how much it matters once realistic deployment factors are allowed in.

\subsection{Advances beyond Teacher Selection}
\label{sec:related-advance}
\looseness=-1
Beyond teacher selection, other work addresses the challenge that student models cannot perfectly replicate teacher reasoning.
Approaches include improving faithfulness of distilled rationales \citep{wang-etal-2023-scott}, curriculum learning \citep{pmlr-v235-feng24e}, augmenting students with external symbolic knowledge bases \citep{liao2025neural}, multi-CoT consistent distillation \citep{chen-etal-2023-mcc}, and self-guided rationale selection \citep{yan-etal-2025-towards}.
These efforts are orthogonal to our focus on teacher selection: they address the separate problem of bridging the imitation gap given a fixed teacher--student pair.

\section{Conclusion}
\label{sec:conclusion}
We revisited the capacity gap in CoT distillation from a practical perspective, focusing on deployment scenarios that genuinely call for distillation: cases where a small, efficient student cannot adequately handle the target task from pretraining alone, such as tasks involving proprietary or domain-specific data with task-specific patterns unavailable from the student's pretraining.
We identified pitfalls in existing evaluation protocols, most notably the lack of comparison against pre-distillation baselines, and proposed a more realistic protocol.
Complementary to prior work that establishes the capacity gap under controlled conditions, our experiments across model families (Qwen2.5 in the main experiments and Gemma-2 in Appendix~\ref{sec:appendix-gemma}) show that its practical impact is not universal and can be outweighed by the overall benefits of stronger teachers, of which increased training data availability is a major component.
Based on these findings, we recommend two conditional guidelines: always verify that distillation improves upon the pre-distillation baseline, and if this is confirmed and candidate teachers differ substantially in performance, prefer the higher-performing teacher.

\section*{Limitations}
\label{sec:limitations}

\paragraph{Benchmark and Model Scope.}
\label{sec:limitations-task}
Our practical evaluation focuses exclusively on BBH as the benchmark where CoT distillation demonstrates genuine improvements over pre-distillation baselines.
While this limits the generalizability of our findings, we note that BBH comprises diverse tasks spanning various reasoning types, and our task selection procedure specifically targeted scenarios where the capacity gap is most likely to manifest.
Regarding model scope, we provide initial cross-family validation using Gemma-2 teachers in Appendix~\ref{sec:appendix-gemma}, but further validation across diverse model families remains important.
We believe BBH represents an appropriate testbed for examining the practical impact of the capacity gap; nevertheless, future work should extend this evaluation to other domains where CoT distillation proves effective, such as code generation and instruction following.

\paragraph{Confounded Factors in Practical Evaluation.}
\label{sec:limitations-filter}
\looseness=-1
By removing cross-teacher data filtering (Section~\ref{sec:re-eval-protocol-filter}), our protocol does not allow us to disentangle the individual contributions of increased data quantity and improved reasoning quality that stronger teachers provide.
However, this is an intentional design choice: our goal is to assess the overall practical impact of using stronger teachers under realistic deployment conditions, rather than to isolate the capacity gap as a phenomenon.
While this approach limits our ability to provide mechanistic explanations, it directly addresses the practitioner's question of which teacher to select given real-world constraints.
We conduct additional experiments with cross-teacher filtering in Appendix~\ref{sec:appendix-filtering}, which confirm that filtering consistently degrades distillation performance, that capacity gap effects remain inconsistent even under equalized data sizes, and that the advantage of stronger teachers is largely driven by increased data \emph{quantity}, more than by differences in CoT \emph{quality}.

\paragraph{Scope of Claims Regarding the Capacity Gap.}
\label{sec:limitations-capacity-gap}
Our findings do not refute the existence of the capacity gap; indeed, both prior work \citep{li-etal-2025-small-models,chen-etal-2025-unveiling-key} and our own results confirm that it manifests in some tasks and settings.
Rather, our contribution is to show that when candidate teachers differ substantially in performance, the benefits of stronger teachers outweigh capacity gap effects.
When candidate teachers perform similarly, capacity gap effects may play a larger role, and other factors such as computational cost or reasoning chain length may guide teacher selection.
Future work should develop more fine-grained selection criteria for such cases.

\paragraph{Statistical Reliability.}
\label{sec:limitations-statistical}
Due to the high computational cost of distillation across configurations, we report single-run results for most configurations, following \citet{li-etal-2025-small-models}.
The BBH test sets are also relatively small (approximately 100 examples per task after splitting), which may amplify noise in per-task results.
To assess this risk, we repeated distillation with three random seeds on four representative tasks (Appendix~\ref{sec:appendix-seeds}); seed averaging leaves our task-level conclusions on these tasks unchanged, and no capacity gap pattern robust to seed variation emerges.
As noted in Section~\ref{sec:re-eval-results-gap}, our capacity gap diagnostic is asymmetric, so noise in marginal cases biases toward inconclusive readings rather than false positives for capacity gap dominance; in addition, our conclusions are drawn from patterns observed across many tasks and settings rather than from individual task-level judgments, further mitigating this limitation.

\paragraph{Task Selection Threshold.}
\label{sec:limitations-threshold}
\looseness=-1
Our task selection criterion (a 30 percentage-point gap between teacher and student ICL performance averages) is not derived from theoretical considerations and may require adjustment for other benchmarks or model families.
Its primary role is to pre-select the tasks a practitioner would realistically consider distilling, relying only on information available before distillation (few-shot ICL performance); serving as a proxy for distillation efficacy is a secondary role.
Accordingly, the threshold does not guarantee that distillation will be effective; indeed, some of the selected tasks turned out to show limited or negative distillation effects (Section~\ref{sec:re-eval-results-efficacy}), and we do not claim that the exact value of 30 is a universally valid efficacy proxy.
Even so, its effectiveness as a proxy is validated beyond the setup used to derive it, across teacher families (Gemma-2), data domains (math), and both sides of the 30-point cutoff (Guideline~1 in Section~\ref{sec:guidelines}), and our conclusions on the capacity gap are themselves unchanged with Gemma-2 teachers (Appendix~\ref{sec:appendix-gemma}) and on either side of the cutoff (Appendix~\ref{sec:appendix-threshold}).
Moreover, the inclusion of both effective and ineffective distillation scenarios reflects the realistic uncertainty practitioners face, providing a practical testbed for examining the capacity gap.

\bibliography{custom}

\clearpage
\appendix

\input{tables/config}
\input{tables/config-lora}

\input{tables/math-data-distill}
\input{tables/math-data-eval}
\input{tables/bbh-data-desc}

\section{Experimental Setup Details}
\label{sec:appendix-setup}
This section provides a detailed description of the experimental setup.
All experiments are conducted on four NVIDIA A100 GPUs, each with 80GB of memory.
Due to the high computational cost of distillation across various configurations, we report results from a single run, following \citet{li-etal-2025-small-models}, except for the multi-seed reliability analysis in Appendix~\ref{sec:appendix-seeds}, which repeats representative configurations with three random seeds.
The datasets used in our experiments are all in English and publicly available for research use.

\subsection{Models}
\label{sec:appendix-setup-models}
We use models from the Qwen2.5 family in our experiments.
For student and teacher models, we use Qwen2.5-\{0.5B, 1.5B, 3B, 7B, 14B, 32B, 72B\}-Instruct, available at \url{https://huggingface.co/collections/Qwen/qwen25}.
For the long-CoT teacher in the short--long setting, we use QwQ-32B-Preview, available at \url{https://huggingface.co/Qwen/QwQ-32B-Preview}.

\subsection{Hyperparameters}
\label{sec:appendix-setup-hyperparameters}
Following the codebase of \citet{li-etal-2025-small-models}, we use \texttt{LLaMA-Factory} \citep{zheng-etal-2024-llamafactory} for fine-tuning student models.
Table~\ref{tab:hyperparameters} summarizes the hyperparameters used in our experiments.
For larger student models (14B and 32B), we employ LoRA \citep{hu2022lora} following \citet{li-etal-2025-small-models}.
The LoRA-specific hyperparameters are provided in Table~\ref{tab:hyperparameters-lora}.

\subsection{MATH Data}
\label{sec:appendix-setup-math}

Table~\ref{tab:training-data} summarizes the distillation datasets used in Section~\ref{sec:pitfall-baseline}.
These datasets were publicly released by \citet{li-etal-2025-small-models}.
They were constructed by generating CoT rationales for the MATH training set \citep{hendrycks2021measuring} and retaining only examples where both teachers in each setting (small--large or short--long) produce correct answers.
Datasets within the same setting have identical sizes due to this cross-teacher filtering (see Section~\ref{sec:pitfall-filter}).

Table~\ref{tab:eval-data} summarizes the evaluation benchmarks used in Section~\ref{sec:pitfall-baseline}.
MATH and GSM8K \citep{cobbe2021training} are standard mathematical reasoning benchmarks; we use their test and validation splits, respectively.
AMC 2023, AIME 2024, and OlympiadBench \citep{he-etal-2024-olympiadbench} are competition-level benchmarks designed to assess advanced mathematical problem-solving abilities.
Following \citet{li-etal-2025-small-models}, we conduct evaluations using the evaluation script from their official implementation along with \texttt{lm-evaluation-harness} \citep{eval-harness}.

\subsection{BBH Data}
\label{sec:appendix-setup-bbh}
We use 15 tasks from BBH \citep{suzgun-etal-2023-challenging,srivastava2023beyond} based on the selection criterion described in Section~\ref{sec:re-eval-protocol-task}.
Table~\ref{tab:bbh-tasks} provides brief descriptions of the selected tasks.
Table~\ref{tab:bbh-pre-all} presents the complete few-shot ICL performance of the pre-distillation models.

Each task contains 250 examples, except for Penguins in a Table, which contains 146 examples.
For each task, we split examples into training and test sets in a 3:2 ratio.
For distillation, we prompt each teacher model to generate CoT rationales for the training examples and retain only those where the teacher produces a correct final answer.
Unlike the MATH experiments, we do not apply cross-teacher filtering; each teacher's distillation dataset consists of all examples that the respective teacher solves correctly, as described in Section~\ref{sec:re-eval-protocol-filter}.
All evaluations are conducted using \texttt{lm-evaluation-harness}, which loads the BBH data and uses few-shot CoT prompts.

\begin{figure}[t]
\centering
\includegraphics[width=1.0\columnwidth,keepaspectratio]{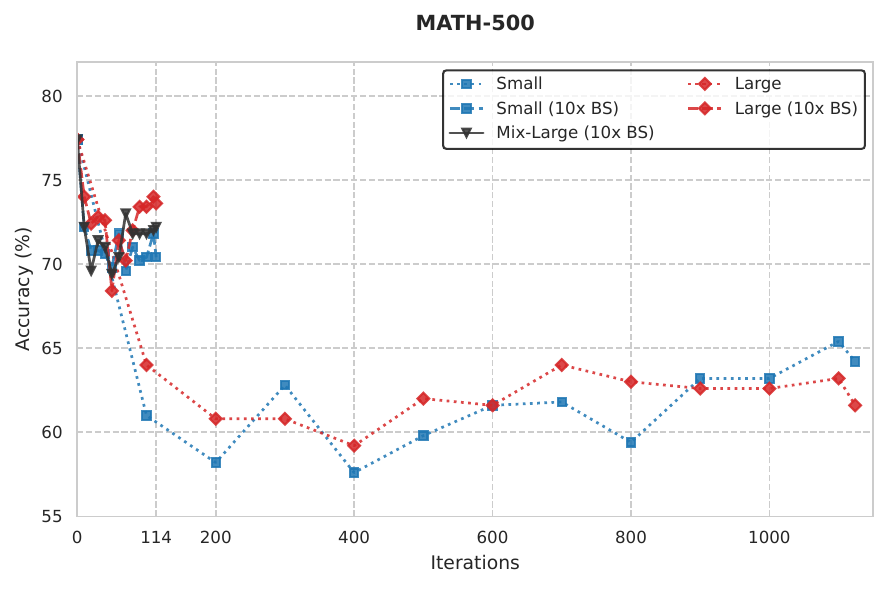}
\caption{
Effect of effective batch size alignment on distillation performance (MATH-500, small--large setting).
(10 $\times$ BS) indicates configurations with 10 times the original effective batch size (batch size $\times$ gradient accumulation steps).
}
\label{fig:math-mix}
\end{figure}

\section{Artifacts from Comparing Degradation}
\label{sec:appendix-artifacts}
As demonstrated in Section~\ref{sec:pitfall-baseline-results}, existing experimental settings exhibit an undesirable situation where distillation \emph{degrades} performance compared to the pre-distillation baseline.
This is problematic not only because it undermines the practical relevance of the experiments, but also because it can lead to misleading evaluations of proposed strategies.
When distillation consistently harms performance, any approach that appears to improve results may simply be mitigating the degree of degradation rather than achieving genuine improvements.
We illustrate this concern with a concrete example from prior work.

\citet{li-etal-2025-small-models} proposed \emph{mix distillation}, which combines rationales from different teachers in a fixed ratio to mitigate the capacity gap.
However, the original implementation uses different hyperparameters from the baseline: mix distillation uses a batch size of 5 with 4 gradient accumulation steps, while standard CoT distillation uses a batch size of 2 with 1 accumulation step.
This discrepancy appears to stem from the mixing ratio: rationales are combined in a 4:1 ratio, requiring larger batches and more accumulation steps to ensure each parameter update reflects rationales from both teachers.
Since the number of epochs is fixed at 2 and other hyperparameters such as learning rate remain the same, mix distillation performs 10 times fewer parameter updates than the baseline.\footnote{Mix distillation performs $\frac{N}{5 \times 4} \times 2 = \frac{N}{10}$ updates, while standard distillation performs $\frac{N}{2 \times 1} \times 2 = N$ updates, where $N$ is the dataset size.}

To isolate the contribution of the data construction strategy, the core component of mix distillation, we aligned the effective batch size (batch size $\times$ gradient accumulation steps) across all methods.
Specifically, we trained students (Qwen2.5-7B-Instruct) using small-teacher and large-teacher distillation data with the same effective batch size as mix distillation (10 $\times$ the original).
We used MATH-500 as the test set and evaluated at fine-grained iteration intervals to examine the effect of the number of update steps.
Figure~\ref{fig:math-mix} presents the results.
When batch size and gradient accumulation are aligned, standard distillation with large-teacher data achieves comparable or better performance than mix distillation: Large (10 $\times$ BS) matches or outperforms Mix-Large (10 $\times$ BS).
This confirms that, in this setting, differences in the number of update steps make it difficult to properly compare distillation strategies.

Beyond this evaluation concern, the analysis also provides an initial probe into the degradation mechanism observed in Section~\ref{sec:pitfall-baseline-results}: configurations with fewer parameter updates, and hence plausibly smaller drift from the initial parameters, consistently suffer less degradation on MATH-500.
This is consistent with an explanation based on optimization-induced drift, i.e., over-updating away from a well-tuned instruction-tuned initialization, although it does not disentangle forgetting, distribution shift, and rationale-style mismatch, which remain open questions.

\input{tables/filtering-avg}
\input{tables/bbh-data-size}

\begin{figure*}[t]
\centering
\includegraphics[width=1.0\textwidth,keepaspectratio]{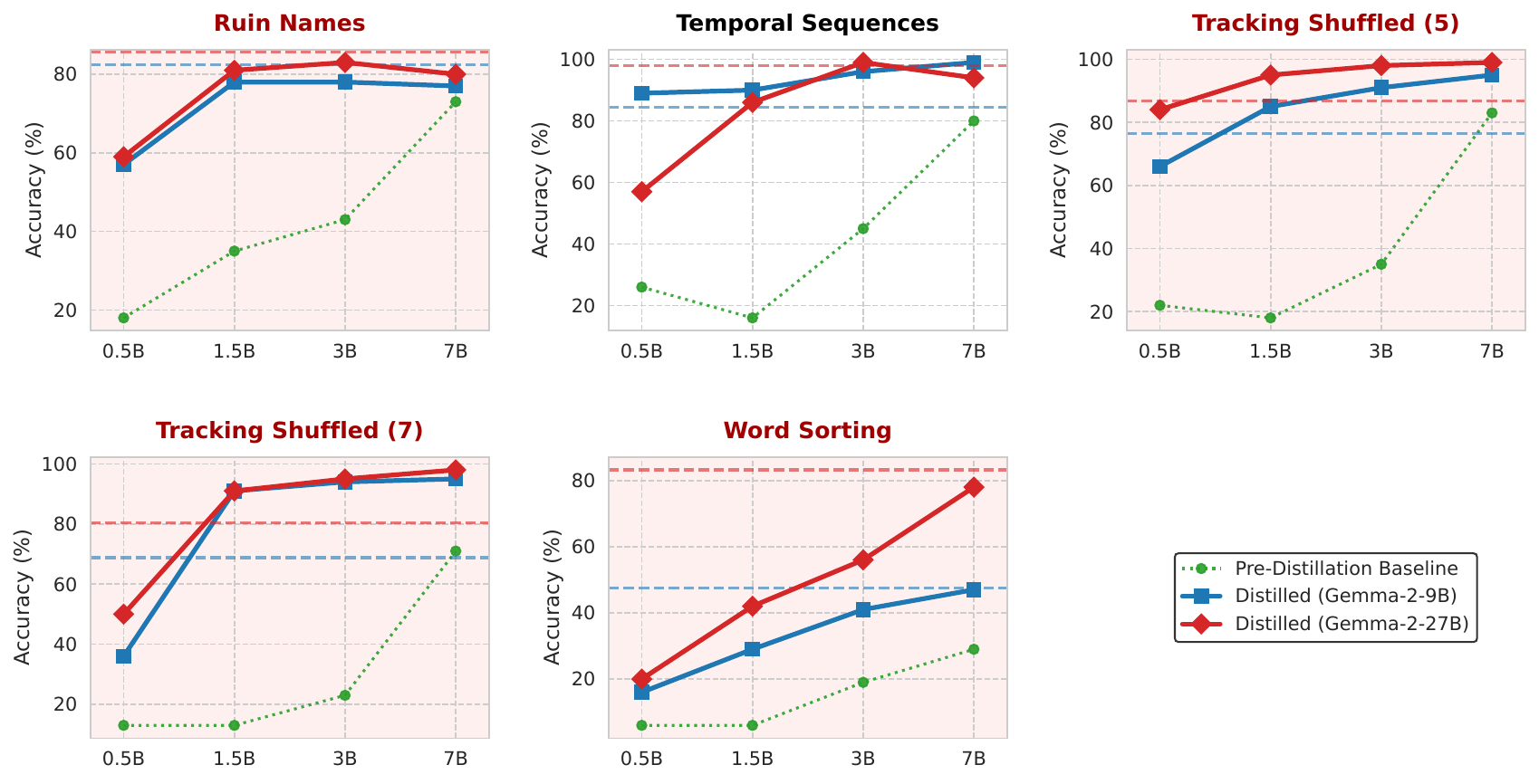}
\caption{
Results on 5 selected BBH tasks with \textbf{Gemma-2} teachers and Qwen2.5 students.
We compare the pre-distillation baseline with students distilled from Gemma-2-9B-it (Small Teacher) and Gemma-2-27B-it (Large Teacher).
Teacher Ceiling indicates the few-shot performance of each teacher model.
Tasks with a \textcolor{textred}{red} background indicate cases where the results did not follow the capacity gap hypothesis.
}
\label{fig:bbh-gemma}
\end{figure*}

\section{Effect of Cross-Teacher Filtering on BBH}
\label{sec:appendix-filtering}

To disentangle the effects of data quantity and reasoning quality, we repeated our BBH experiments (Section~\ref{sec:re-eval}) with the cross-teacher filtering protocol used in prior work, where training data is restricted to examples that both teachers solve correctly.

Table~\ref{tab:filtering-avg} reports the average accuracy across all 15 BBH tasks (averaged over four student sizes: 0.5B, 1.5B, 3B, 7B).
Cross-teacher filtering consistently degrades average accuracy compared to the unfiltered setting.
Notably, the degradation is asymmetric in the small--large setting (Large: $-3.61$ vs.\ Small: $-1.78$).
The data reduction is likewise asymmetric: as shown in the bottom row of Table~\ref{tab:bbh-data-size}, filtering shrinks the large teacher's data from 1,928 to 1,684 examples in total ($-12.7\%$) but the small teacher's only from 1,769 to 1,684 ($-4.8\%$).
Filtering thus ends up disproportionately penalizing the stronger teacher, stripping it of its additional correct examples.
In the short--long setting, no such asymmetry appears, either in the degradation (Short: $-3.16$ vs.\ Long: $-2.97$) or in the total data size (1,919 vs.\ 1,924 examples).
However, since the two teachers solve different examples, filtering still removes a similarly large amount of data from both (down to 1,790 examples; $-6.7\%$ and $-7.0\%$), and both degrade to a similarly large extent.
Across the four teachers, the size of the degradation thus corresponds to the amount of data lost to filtering (Small: $-4.8\%$ data, $-1.78$ points; Short and Long: about $-7\%$, $-3.16$ and $-2.97$; Large: $-12.7\%$, $-3.61$), indicating that the degradation under filtering is driven by the loss of training data.

Per-task accuracies behind these averages are provided in Tables~\ref{tab:unfiltered-small-large}--\ref{tab:filtering-short-long}.
The effect of filtering is most visible in Dyck Languages and Word Sorting, where the stronger teacher provided substantially more data in the unfiltered setting (approximately 1.3--1.8$\times$ more examples; Table~\ref{tab:bbh-data-size}).
Without filtering, the stronger teacher consistently produced better or comparable students regardless of student size in these tasks.
Equalizing data sizes through filtering substantially weakened this pattern.
Specifically, in Dyck Languages, the average absolute accuracy gap between students of the two teachers (across student sizes) shrank from $18.50$ points (unfiltered) to $3.25$ points (filtered) in the small--large setting, and from $15.50$ to $5.75$ points in the short--long setting.
In Word Sorting, the gap shrank from $5.00$ to $1.75$ points in the small--large setting, and from $11.50$ to $4.75$ points in the short--long setting.
These results provide direct evidence that the advantage of stronger teachers is largely driven by increased data \emph{quantity}, more than by differences in CoT \emph{quality} between the two teachers.

\input{tables/band-tasks}

\section{Generalization to a Different Model Family}
\label{sec:appendix-gemma}

Our main experiments use the Qwen2.5 family for both teachers and students.
To examine whether our findings generalize beyond a single model family, we conduct additional experiments using Gemma-2-\{9B, 27B\}-it \citep{gemmateam2024gemma2} as teachers while keeping the same Qwen2.5 students.
This cross-family setting follows \citet{li-etal-2025-small-models}, who also employed Gemma-2-9B-it and Gemma-2-27B-it as small and large teachers in their small--large setting.
As in \citet{li-etal-2025-small-models}, we evaluate only the small--large setting since the Gemma-2 family does not include a long-CoT teacher.

Applying the same 30 percentage-point gap threshold used in Section~\ref{sec:re-eval-protocol-task}, we select 5 tasks for distillation: Ruin Names, Temporal Sequences, Tracking Shuffled Objects (5), Tracking Shuffled Objects (7), and Word Sorting.
The pre-distillation performance of all models is reported in Table~\ref{tab:bbh-pre-all-gemma}.

Figure~\ref{fig:bbh-gemma} presents the results.
Among the 5 selected tasks, distillation is effective in all of them, consistently improving over the pre-distillation baseline.
Regarding capacity gap effects, only 1 out of 5 tasks (Temporal Sequences) exhibits such effects, where the smaller teacher (9B) yields better students at smaller student sizes.
In the remaining 4 tasks, the larger teacher (27B) consistently produces better or comparable students across all student sizes.
These results are consistent with our main findings from the Qwen-only experiments (Section~\ref{sec:re-eval}), confirming that the impact of capacity gap effects does not consistently dominate across tasks, even when a different teacher model family is used.

\section{Validation of the Task-Selection Threshold}
\label{sec:appendix-threshold}

Our practical protocol selects BBH tasks whose ICL gap between the teacher and student group averages exceeds 30 percentage points (Section~\ref{sec:re-eval-protocol-task}).
This appendix validates the threshold on both sides of the cutoff: from below, by newly distilling three tasks just under the threshold, and from above, by re-reading our results under stricter cutoffs.

\paragraph{Distillation below the Threshold.}
We newly distill three BBH tasks from the 20--30 point band just below the cutoff (Table~\ref{tab:bbh-pre-all}): Multistep Arithmetic Two (ICL gap 28.8), Object Counting (28.9), and Disambiguation QA (26.4).
The protocol is identical to that of the main BBH experiments (Qwen2.5 students, the same four teachers, no cross-teacher filtering).
Table~\ref{tab:band-tasks} reports the results, which reinforce the threshold in two respects.

First, distillation is not reliably effective on any of the three tasks.
Every one of them has configurations where the distilled student falls below the pre-distillation baseline, so none of them meets the efficacy criterion of consistent improvement in either teacher setting.
Even the closest case fails: on Disambiguation QA in the short--long setting, the long teacher improves at every student size, but the short teacher's 7B student only ties its baseline of 75.0.
None of the tested below-threshold tasks would thus qualify as distillation-effective, suggesting that the threshold reflects a real change in distillation reliability rather than an arbitrary single point.

Second, our capacity gap conclusions also hold below the cutoff.
On Multistep Arithmetic Two, the \emph{more capable} teachers produce the best 0.5B students (89.0 vs.\ 58.0 for large vs.\ small; 83.0 vs.\ 62.0 for long vs.\ short), the opposite of capacity gap dominance.
On Object Counting, the small--large winner alternates with student size.
Only Disambiguation QA in the small--large setting shows a capacity gap pattern, and it does not carry over to the short--long setting.
As Table~\ref{tab:bbh-pre-all} shows, the teacher ICL accuracies on these three tasks are within 5.2 points of each other in each setting, so the data-quantity advantage that drives Dyck Languages and Word Sorting (Section~\ref{sec:guidelines-2}) does not apply here; this matches our analysis of ineffective distillation tasks (Section~\ref{sec:re-eval-results-ineffective}).

\paragraph{Stricter Cutoffs.}
Since ICL gaps are already measured for all tasks (Table~\ref{tab:bbh-pre-all}), stricter cutoffs can be checked on our existing results without new runs.
Table~\ref{tab:threshold-sensitivity} summarizes the outcome for cutoffs of 30, 35, and 40 points.
Most selected tasks remain distillation-effective under the stricter cutoffs (8 of 10 at $>$35 and 6 of 8 at $>$40, in both teacher settings), so the proxy keeps identifying suitable tasks.
The capacity gap pattern also stays inconsistent across settings at every cutoff, and the two tasks where the stronger teacher's data-quantity advantage drives the result (Dyck Languages and Word Sorting) remain included even at the strictest cutoff.
Our conclusions therefore do not hinge on the exact value of 30; the threshold is an operational heuristic whose primary role is to pre-select the tasks a practitioner would realistically consider distilling (see the \hyperref[sec:limitations-threshold]{Limitations} section).

\input{tables/threshold-sensitivity}
\input{tables/seeds}

\section{Multi-Seed Reliability Analysis}
\label{sec:appendix-seeds}

Our main results are reported from a single run per configuration due to the high computational cost of distillation.
To assess seed-level variability, we repeat distillation with three random seeds on four representative tasks.
Because the findings most in need of verification are those that run counter to expectations from prior work, namely that distillation can fail to improve the student and that capacity gap patterns can be absent, we specifically select tasks carrying these readings: Formal Fallacies and Penguins in a Table, where distillation showed limited efficacy in the single-run results (Section~\ref{sec:re-eval-results-ineffective}), and Date Understanding and Dyck Languages, distillation-effective tasks where no capacity gap pattern was observed (Section~\ref{sec:re-eval-results-gap}).
The runs cover the three smallest students (0.5B, 1.5B, and 3B) and both teacher settings.
Table~\ref{tab:seeds} reports the mean and standard deviation over the three seeds.

The seed-averaged results reveal no capacity gap pattern robust to seed variation.
Our diagnostic pattern for capacity gap effects requires the less capable teacher (small or short) to \emph{actively} outperform the more capable one (large or long) at the smallest student sizes, with the advantage diminishing as the student grows (Section~\ref{sec:re-eval-results-gap}).
Across the 24 comparisons in Table~\ref{tab:seeds} (four tasks, three student sizes, and two settings), leads of the less capable teacher are, with one exception discussed below, at most 2.8 points (Penguins in a Table at 3B, $80.2 \pm 5.4$ vs.\ $77.4 \pm 4.3$) and stay within the seed-level variation.
The exception is Dyck Languages in the short--long setting, where the short teacher leads at every student size.
This lead, however, does not fit the capacity gap signature: it jumps from 4.3 points at 0.5B to about 20 points at the two larger sizes rather than diminishing, and the short teacher is the \emph{stronger} teacher on this task (few-shot accuracy 58.8 vs.\ 40.4), so the lead is consistent with the data-quantity explanation (Section~\ref{sec:guidelines-2}; Appendix~\ref{sec:appendix-filtering}).
In contrast, the advantage of a substantially stronger teacher (Section~\ref{sec:re-eval-results-practical}) remains clearly visible above the seed-level variation.
Among the four tasks, Dyck Languages is the only one where the teachers differ substantially in performance, and its stronger teacher leads well beyond the seed variation in both settings: the short teacher in the short--long setting, as discussed above, and the large teacher in the small--large setting (few-shot accuracy 61.2 vs.\ 31.6), which leads by 6.4 points at 0.5B and by 36.7 points at 1.5B ($49.7 \pm 2.1$ vs.\ $13.0 \pm 1.7$), the latter an order of magnitude above the seed variation.
The multi-seed results therefore provide no support for capacity gap dominance on any of the four tasks.

Beyond the capacity gap pattern, seed averaging also preserves the distillation-efficacy readings: Date Understanding and Dyck Languages remain distillation-effective, while Formal Fallacies and Penguins in a Table remain tasks with limited distillation efficacy (underlined scores in Table~\ref{tab:seeds}).
The task-level conclusions drawn in the paper thus remain unchanged for all four tasks.

\input{tables/fixed-budget}

\section{Evaluation under a Fixed Inference Budget}
\label{sec:appendix-budget}

Practitioners who prefer short reasoning to save inference costs may cap the generation length at deployment, in which case a student can be cut off in the middle of its reasoning.
To examine how capacity gap effects behave under such a constraint, we re-evaluate the BBH students from our main experiments (Section~\ref{sec:re-eval}) under fixed generation budgets, capping the maximum generation length at $N$ tokens and scoring with each task's original evaluation configuration.
Table~\ref{tab:fixed-budget} reports macro-average accuracy over the 15 tasks for the students distilled from the short and long teachers, together with the mean number of tokens actually generated at each budget, i.e., the effective inference cost.

A tight budget can indeed make the short teacher preferable, but not consistently.
At $N{=}128$ to $512$, the short teacher leads for the 1.5B, 3B, and 7B students (e.g., 81.0 vs.\ 78.7 for the 7B student at $N{=}512$).
However, at 0.5B, the long teacher wins at every budget, including $N{=}128$, where truncation nearly equalizes the tokens consumed across models; at 7B, the short teacher stays ahead even at $N{=}4000$, where truncation is rare and unlikely to explain the advantage; and at 1.5B and 3B, the ranking returns to the long teacher by $N{=}1024$, with differences within one point.
A limited inference budget therefore does not uniformly favor the short-CoT teacher: the preferable teacher varies with student size and budget, consistent with our central claim that capacity gap effects do not consistently dominate.

\section{Use of AI Assistants}
\label{sec:appendix-ai-use}
AI assistants were used for translation, improving the fluency of the text, and aggregating the experimental results into tables and figures.
All outputs were reviewed and verified by the authors.

\input{tables/unfiltered-small-large}
\input{tables/unfiltered-short-long}

\input{tables/filtering-small-large}
\input{tables/filtering-short-long}

\input{tables/bbh-pre-all}
\input{tables/bbh-pre-all-gemma}

\end{document}

%% file: tables/config.tex
\begin{table}[t]
\centering
\adjustbox{max width=\columnwidth}{
\begin{tabular}{ll}
\toprule
\textbf{Hyperparameter} & \textbf{Value} \\
\midrule
\multicolumn{2}{l}{\textit{Method}} \\
Stage & SFT \\
Finetuning type & Full \\
DeepSpeed config & ds\_z3\_config.json \\
Flash attention & FA2 \\
\midrule
\multicolumn{2}{l}{\textit{Dataset}} \\
Template & qwen \\
Max samples & 10,000 \\
Cutoff length & 8,192 \\
Preprocessing workers & 32 \\
\midrule
\multicolumn{2}{l}{\textit{Training}} \\
Batch size (per device) & 2 \\
Gradient accumulation steps & 1 \\
Learning rate & 1.0e-5 \\
Number of epochs & 2 \\
LR scheduler & Cosine \\
Precision & BF16 \\
\midrule
\multicolumn{2}{l}{\textit{Evaluation}} \\
Validation size & 0.01 \\
Eval batch size (per device) & 1 \\
Eval strategy & Steps \\
Eval steps & 100 \\
\bottomrule
\end{tabular}
}
\caption{Hyperparameters for full fine-tuning of smaller student models.}
\label{tab:hyperparameters}
\end{table}

%% file: tables/config-lora.tex
\begin{table}[t]
\centering
\adjustbox{max width=\columnwidth}{
\begin{tabular}{ll}
\toprule
\textbf{Hyperparameter} & \textbf{Value} \\
\midrule
\multicolumn{2}{l}{\textit{Method}} \\
Stage & SFT \\
Finetuning type & LoRA \\
Lora Target & All \\
DeepSpeed config & ds\_z3\_config.json \\
Flash attention & FA2 \\
\midrule
\multicolumn{2}{l}{\textit{Dataset}} \\
Template & qwen \\
Max samples & 10,000 \\
Cutoff length & 8,192 \\
Preprocessing workers & 32 \\
\midrule
\multicolumn{2}{l}{\textit{Training}} \\
Batch size (per device) & 1 \\
Gradient accumulation steps & 1 \\
Learning rate & 1.0e-4 \\
Number of epochs & 2 \\
LR scheduler & Cosine \\
Precision & BF16 \\
Warmup ratio & 0.03 \\
\midrule
\multicolumn{2}{l}{\textit{Evaluation}} \\
Validation size & 0.01 \\
Eval batch size (per device) & 1 \\
Eval strategy & Steps \\
Eval steps & 100 \\
\bottomrule
\end{tabular}
}
\caption{Hyperparameters for LoRA fine-tuning of larger student models (14B and 32B).}
\label{tab:hyperparameters-lora}
\end{table}

%% file: tables/math-data-distill.tex
\begin{table*}[t]
\centering
\adjustbox{max width=\textwidth}{
\begin{tabular}{lrp{\textwidth}}
\toprule
\textbf{Distillation Data} & \textbf{Size} & \textbf{Source} \\
\midrule
MATH w/ Small Teacher & 4,539 & \url{https://huggingface.co/datasets/UWNSL/MATH_training_split_distill_small_teacher} \\
MATH w/ Large Teacher & 4,539 & \url{https://huggingface.co/datasets/UWNSL/MATH_training_split_distill_large_teacher} \\
MATH w/ Short Teacher & 5,383 & \url{https://huggingface.co/datasets/UWNSL/MATH_training_split_short_cot} \\
MATH w/ Long Teacher & 5,383 & \url{https://huggingface.co/datasets/UWNSL/MATH_training_split_long_cot} \\
\bottomrule
\end{tabular}
}
\caption{Distillation datasets for MATH experiments in Section~\ref{sec:pitfall-baseline}.}
\label{tab:training-data}
\end{table*}

%% file: tables/math-data-eval.tex
\begin{table*}[t]
\centering
\adjustbox{max width=\textwidth}{
\begin{tabular}{lrp{\textwidth}}
\toprule
\textbf{Evaluation Data} & \textbf{Size} & \textbf{Source} \\
\midrule
MATH & 5,000 & \url{https://huggingface.co/datasets/EleutherAI/hendrycks_math} \\
GSM8K & 1,319 & \url{https://huggingface.co/datasets/openai/gsm8k} \\
AMC 2023 & 40 & \url{https://github.com/Small-Model-Gap/Small-Model-Learnability-Gap/tree/main/lm-evaluation-harness/math_eval_data/amc23} \\
AIME 2024 & 30 & \url{https://github.com/Small-Model-Gap/Small-Model-Learnability-Gap/tree/main/lm-evaluation-harness/math_eval_data/aime24} \\
OlympiadBench & 675 & \url{https://github.com/Small-Model-Gap/Small-Model-Learnability-Gap/tree/main/lm-evaluation-harness/math_eval_data/olympiadbench} \\
\bottomrule
\end{tabular}
}
\caption{Evaluation benchmarks for MATH experiments in Section~\ref{sec:pitfall-baseline}.}
\label{tab:eval-data}
\end{table*}

%% file: tables/bbh-data-desc.tex
\begin{table*}[t]
\centering
\adjustbox{max width=\textwidth}{
\begin{tabular}{lp{\textwidth}}
\toprule
\textbf{Task} & \textbf{Description} \\
\midrule
Date Understanding & Read dates from context and answer questions involving format conversion or duration calculation. \\
Dyck Languages & Complete a parenthesis sequence by supplying the missing closing brackets. \\
Formal Fallacies & Determine the logical validity of an informal argument. \\
Logical Deduction \{3, 5, 7\} & Given clues about the relative positions of 3, 5, or 7 objects, identify the correct position of a specified object. \\
Penguins in a Table & Answer questions about specific penguins or their counts based on a table of penguin attributes. \\
Reasoning Colored Objects & Given a description of object arrangements, identify the color of a specified object. \\
Ruin Names & Alter exactly one character of a given artist, band, or movie name to produce a humorous variant. \\
Salient Translation Error & Compare a German source sentence with an erroneous English translation and classify the type of translation error such as proper nouns, numbers, or negation. \\
Temporal Sequences & Determine the time period during which a specific event occurred, given a person's daily schedule. \\
Tracking Shuffled Objects \{3, 5, 7\} & Track a series of swap operations on 3, 5, or 7 objects and identify the final owner of each object. \\
Word Sorting & Sort a list of words into alphabetical order. \\
\bottomrule
\end{tabular}
}
\caption{Descriptions of the 15 BBH tasks used in our experiments.}
\label{tab:bbh-tasks}
\end{table*}

%% file: tables/filtering-avg.tex
\begin{table}[t]
\centering
\begin{tabular}{lccc}
\toprule
Teacher & No Filter & Filtered & Diff. \\
\midrule
Small & 67.49 & 65.70 & $-1.78$ \\
Large & 67.95 & 64.34 & $-3.61$ \\
\midrule
Short & 68.09 & 64.92 & $-3.16$ \\
Long & 69.09 & 66.11 & $-2.97$ \\
\bottomrule
\end{tabular}
\caption{Average accuracy across 15 BBH tasks (averaged over four student sizes) \textbf{with and without cross-teacher filtering}. Diff.\ = Filtered $-$ No Filter.}
\label{tab:filtering-avg}
\end{table}

%% file: tables/bbh-data-size.tex
\begin{table*}[t]
\centering
\adjustbox{max width=\textwidth}{
\begin{tabular}{lcccccccc}
\toprule
& \multicolumn{4}{c}{\textbf{Small--Large}} & \multicolumn{4}{c}{\textbf{Short--Long}} \\
\cmidrule(lr){2-5} \cmidrule(lr){6-9}
\textbf{Task} & \textbf{Small} & \textbf{Large} & \textbf{Ratio} & \textbf{Filtered} & \textbf{Short} & \textbf{Long} & \textbf{Ratio} & \textbf{Filtered} \\
\midrule
Date Understanding & 136 & 138 & 1.01 & 129 & 136 & 135 & 1.01 & 130 \\
Dyck Languages & 49 & 90 & \textbf{1.84} & 44 & 88 & 61 & \textbf{1.44} & 47 \\
Formal Fallacies & 126 & 137 & 1.09 & 123 & 135 & 150 & 1.11 & 135 \\
Logical Deduction (3) & 148 & 147 & 1.01 & 146 & 148 & 146 & 1.01 & 146 \\
Logical Deduction (5) & 123 & 137 & 1.11 & 113 & 130 & 138 & 1.06 & 120 \\
Logical Deduction (7) & 116 & 125 & 1.08 & 99 & 126 & 125 & 1.01 & 106 \\
Penguins in a Table & 84 & 86 & 1.02 & 84 & 86 & 84 & 1.02 & 83 \\
Reasoning about Colored Objects & 139 & 148 & 1.06 & 137 & 149 & 150 & 1.01 & 149 \\
Ruin Names & 130 & 133 & 1.02 & 123 & 125 & 124 & 1.01 & 115 \\
Salient Translation Error Detection & 99 & 93 & 1.06 & 81 & 105 & 97 & 1.08 & 89 \\
Temporal Sequences & 144 & 150 & 1.04 & 144 & 150 & 146 & 1.03 & 146 \\
Tracking Shuffled Objects (3) & 135 & 150 & 1.11 & 135 & 149 & 149 & 1.00 & 148 \\
Tracking Shuffled Objects (5) & 132 & 147 & 1.11 & 130 & 150 & 146 & 1.03 & 146 \\
Tracking Shuffled Objects (7) & 132 & 146 & 1.11 & 128 & 149 & 146 & 1.02 & 145 \\
Word Sorting & 76 & 101 & \underline{1.33} & 68 & 93 & 127 & \underline{1.37} & 85 \\
\midrule
Total & 1,769 & 1,928 & & 1,684 & 1,919 & 1,924 & & 1,790 \\
\bottomrule
\end{tabular}
}
\caption{Training data sizes for the BBH distillation experiments. For each teacher pairing, we report the number of training examples that each teacher solves correctly, which constitutes its distillation dataset in the unfiltered setting; the ratio between these two numbers, dividing whichever is greater by the other (\textbf{Ratio}); and the number of examples that both teachers solve correctly, which is used under cross-teacher filtering (\textbf{Filtered}). In each setting, the highest ratio among the tasks is shown in \textbf{bold} and the second highest is \underline{underlined}. The bottom row reports the total data sizes over the 15 tasks.}
\label{tab:bbh-data-size}
\end{table*}

%% file: tables/band-tasks.tex
\begin{table*}[t]
\centering
\begin{tabular}{llccccc}
\toprule
\textbf{Task (ICL Gap)} & \textbf{Student} & \textbf{Baseline} & \textbf{Small} & \textbf{Large} & \textbf{Short} & \textbf{Long} \\
\midrule
\multirow{4}{*}{Multistep Arithmetic Two (28.8)} & 0.5B & 27.0 & 58.0 & 89.0 & 62.0 & 83.0 \\
 & 1.5B & 67.0 & 83.0 & \underline{65.0} & 84.0 & 93.0 \\
 & 3B & 90.0 & \underline{83.0} & \underline{89.0} & 94.0 & 98.0 \\
 & 7B & 95.0 & \underline{73.0} & \underline{88.0} & 96.0 & \underline{82.0} \\
\midrule
\multirow{4}{*}{Object Counting (28.9)} & 0.5B & 48.0 & 53.0 & \underline{40.0} & \underline{43.0} & \underline{26.0} \\
 & 1.5B & 56.0 & \underline{50.0} & \underline{53.0} & \underline{51.0} & \underline{44.0} \\
 & 3B & 62.0 & \underline{52.0} & \underline{45.0} & \underline{51.0} & \underline{48.0} \\
 & 7B & 82.0 & \underline{55.0} & \underline{56.0} & \underline{53.0} & \underline{40.0} \\
\midrule
\multirow{4}{*}{Disambiguation QA (26.4)} & 0.5B & 29.0 & 53.0 & 41.0 & 55.0 & 34.0 \\
 & 1.5B & 51.0 & 71.0 & \underline{35.0} & 55.0 & 58.0 \\
 & 3B & 61.0 & 79.0 & \underline{59.0} & 79.0 & 66.0 \\
 & 7B & 75.0 & 87.0 & 88.0 & 75.0 & 82.0 \\
\bottomrule
\end{tabular}
\caption{Accuracy on the test split for three BBH tasks whose ICL gap falls just \textbf{below the 30-point threshold} (in the 20--30 point band), distilled under the same protocol as the main experiments. Baseline is the pre-distillation few-shot accuracy; Small/Large (Qwen2.5-14B/72B-Instruct) and Short/Long (Qwen2.5-32B-Instruct/QwQ-32B-Preview) denote students distilled from the respective teachers. \underline{Underlined} scores fall below the pre-distillation baseline.}
\label{tab:band-tasks}
\end{table*}

%% file: tables/threshold-sensitivity.tex
\begin{table}[t]
\centering
\adjustbox{max width=\columnwidth}{
\begin{tabular}{cccccc}
\toprule
& & \multicolumn{2}{c}{\textbf{Small--Large}} & \multicolumn{2}{c}{\textbf{Short--Long}} \\
\cmidrule(lr){3-4} \cmidrule(lr){5-6}
\textbf{Cutoff} & \textbf{\#Tasks} & \textbf{Effective} & \textbf{Gap-Cons.} & \textbf{Effective} & \textbf{Gap-Cons.} \\
\midrule
$>$30 & 15 & 11 & 8 & 10 & 2 \\
$>$35 & 10 & 8 & 6 & 8 & 2 \\
$>$40 & 8 & 6 & 4 & 6 & 2 \\
\bottomrule
\end{tabular}
}
\caption{Sensitivity of our findings to the task-selection cutoff. \textbf{\#Tasks} counts the BBH tasks whose ICL gap (Table~\ref{tab:bbh-pre-all}) exceeds the cutoff; for each teacher setting, \textbf{Effective} counts the selected tasks where distillation consistently improves over the pre-distillation baseline, and \textbf{Gap-Cons.} counts the effective tasks that exhibit the capacity-gap diagnostic pattern (Section~\ref{sec:re-eval-results-gap}). The $>$30 row is the setting used in the main experiments.}
\label{tab:threshold-sensitivity}
\end{table}

%% file: tables/seeds.tex
\begin{table*}[t]
\centering
\adjustbox{max width=\textwidth}{
\begin{tabular}{llccccc}
\toprule
\textbf{Task} & \textbf{Student} & \textbf{Baseline} & \textbf{Small} & \textbf{Large} & \textbf{Short} & \textbf{Long} \\
\midrule
\multirow{3}{*}{Dyck Languages} & 0.5B & 1.0 & 4.3 $\pm$ 2.9 & 10.7 $\pm$ 3.1 & 10.3 $\pm$ 1.2 & 6.0 $\pm$ 1.0 \\
 & 1.5B & 1.0 & 13.0 $\pm$ 1.7 & 49.7 $\pm$ 2.1 & 40.0 $\pm$ 2.6 & 19.0 $\pm$ 6.0 \\
 & 3B & 2.0 & 36.7 $\pm$ 11.2 & 44.7 $\pm$ 11.5 & 53.7 $\pm$ 7.8 & 33.7 $\pm$ 8.4 \\
\midrule
\multirow{3}{*}{Date Understanding} & 0.5B & 36.0 & 49.3 $\pm$ 3.5 & 57.0 $\pm$ 1.7 & 60.0 $\pm$ 5.6 & 64.3 $\pm$ 3.1 \\
 & 1.5B & 53.0 & 85.3 $\pm$ 1.2 & 86.0 $\pm$ 1.0 & 81.0 $\pm$ 1.7 & 82.0 $\pm$ 1.7 \\
 & 3B & 67.0 & 87.3 $\pm$ 3.5 & 87.0 $\pm$ 1.0 & 85.0 $\pm$ 5.2 & 84.0 $\pm$ 3.6 \\
\midrule
\multirow{3}{*}{Formal Fallacies} & 0.5B & 53.0 & \underline{51.7 $\pm$ 2.9} & \underline{50.7 $\pm$ 3.1} & \underline{51.7 $\pm$ 1.2} & \underline{49.3 $\pm$ 6.1} \\
 & 1.5B & 53.0 & 57.0 $\pm$ 4.6 & 61.3 $\pm$ 2.1 & 54.7 $\pm$ 3.1 & 59.7 $\pm$ 5.5 \\
 & 3B & 63.0 & 65.3 $\pm$ 4.2 & 71.7 $\pm$ 2.5 & 63.0 $\pm$ 3.5 & 73.3 $\pm$ 1.5 \\
\midrule
\multirow{3}{*}{Penguins in a Table} & 0.5B & 32.2 & 39.5 $\pm$ 7.6 & 37.3 $\pm$ 10.3 & 40.1 $\pm$ 12.7 & 42.4 $\pm$ 2.9 \\
 & 1.5B & 64.4 & 68.4 $\pm$ 2.6 & 68.9 $\pm$ 8.4 & 65.5 $\pm$ 2.0 & 70.1 $\pm$ 4.3 \\
 & 3B & 81.4 & \underline{80.2 $\pm$ 5.4} & \underline{77.4 $\pm$ 4.3} & 81.9 $\pm$ 1.0 & 84.2 $\pm$ 8.4 \\
\bottomrule
\end{tabular}
}
\caption{\textbf{Multi-seed} distillation results on four representative BBH tasks. Accuracy on the test split is reported as mean $\pm$ standard deviation over three random seeds; Baseline is the pre-distillation few-shot accuracy on the same test split. \underline{Underlined} scores indicate that the mean falls below the pre-distillation baseline.}
\label{tab:seeds}
\end{table*}

%% file: tables/fixed-budget.tex
\begin{table*}[t]
\centering
\adjustbox{max width=\textwidth}{
\begin{tabular}{lcccc}
\toprule
\textbf{Student} & $\bm{N{=}128}$ & $\bm{N{=}512}$ & $\bm{N{=}1024}$ & $\bm{N{=}4000}$ \\
\midrule
0.5B & 32.7 (115) / \textbf{33.7} (115) & 46.8 (194) / \textbf{48.5} (213) & 46.9 (211) / \textbf{49.1} (251) & 46.9 (299) / \textbf{49.1} (449) \\
1.5B & \textbf{44.1} (114) / 43.1 (116) & \textbf{67.3} (193) / 66.6 (209) & 67.4 (210) / \textbf{68.0} (240) & 67.4 (284) / \textbf{68.2} (339) \\
3B & \textbf{48.3} (114) / 46.7 (116) & \textbf{75.7} (193) / 74.3 (222) & 75.9 (207) / \textbf{76.5} (267) & 75.9 (244) / \textbf{76.6} (448) \\
7B & \textbf{50.1} (114) / 47.5 (116) & \textbf{81.0} (198) / 78.7 (226) & \textbf{82.7} (215) / 82.2 (269) & \textbf{83.1} (235) / 82.6 (413) \\
\bottomrule
\end{tabular}
}
\caption{Macro-average accuracy over the 15 BBH tasks under a \textbf{fixed inference budget} of $N$ generated tokens in the \emph{short--long} setting, obtained by re-evaluating the students with the maximum generation length capped at $N$ tokens. Each cell shows the accuracy of the short teacher's / long teacher's student, with the mean number of tokens actually generated at that budget in parentheses; \textbf{bold} marks the better of the pair.}
\label{tab:fixed-budget}
\end{table*}

%% file: tables/unfiltered-small-large.tex
\begin{table*}[t]
\centering
\adjustbox{max width=\textwidth}{
\begin{tabular}{lcccccccc}
\toprule
& \multicolumn{4}{c}{\textbf{Small Teacher (14B)}} & \multicolumn{4}{c}{\textbf{Large Teacher (72B)}} \\
\cmidrule(lr){2-5} \cmidrule(lr){6-9}
\textbf{Task} & \textbf{0.5B} & \textbf{1.5B} & \textbf{3B} & \textbf{7B} & \textbf{0.5B} & \textbf{1.5B} & \textbf{3B} & \textbf{7B} \\
\midrule
Date Understanding & 50.0 & 87.0 & 90.0 & 91.0 & 56.0 & 87.0 & 89.0 & 95.0 \\
Dyck Languages & 6.0 & 14.0 & 26.0 & 36.0 & 11.0 & 49.0 & 56.0 & 40.0 \\
Formal Fallacies & 58.0 & 63.0 & 59.0 & 68.0 & 51.0 & 56.0 & 74.0 & 76.0 \\
Logical Deduction (3) & 52.0 & 87.0 & 96.0 & 93.0 & 50.0 & 84.0 & 96.0 & 93.0 \\
Logical Deduction (5) & 29.0 & 36.0 & 59.0 & 62.0 & 21.0 & 57.0 & 61.0 & 75.0 \\
Logical Deduction (7) & 19.0 & 45.0 & 54.0 & 58.0 & 18.0 & 43.0 & 42.0 & 53.0 \\
Penguins in a Table & 45.8 & 66.1 & 74.6 & 84.7 & 49.2 & 74.6 & 79.7 & 79.7 \\
Reasoning about Colored Objects & 41.0 & 77.0 & 83.0 & 92.0 & 34.0 & 77.0 & 85.0 & 90.0 \\
Ruin Names & 66.0 & 80.0 & 82.0 & 83.0 & 58.0 & 79.0 & 74.0 & 85.0 \\
Salient Translation Error Detection & 49.0 & 59.0 & 67.0 & 73.0 & 34.0 & 42.0 & 61.0 & 61.0 \\
Temporal Sequences & 78.0 & 85.0 & 97.0 & 100.0 & 62.0 & 70.0 & 98.0 & 99.0 \\
Tracking Shuffled Objects (3) & 95.0 & 100.0 & 97.0 & 100.0 & 69.0 & 99.0 & 98.0 & 100.0 \\
Tracking Shuffled Objects (5) & 99.0 & 85.0 & 99.0 & 98.0 & 94.0 & 99.0 & 100.0 & 100.0 \\
Tracking Shuffled Objects (7) & 39.0 & 94.0 & 99.0 & 96.0 & 56.0 & 97.0 & 100.0 & 99.0 \\
Word Sorting & 18.0 & 30.0 & 37.0 & 42.0 & 15.0 & 30.0 & 39.0 & 57.0 \\
\bottomrule
\end{tabular}
}
\caption{Per-task accuracy on BBH \textbf{without cross-teacher filtering} in the \emph{small--large} setting, i.e., our main experiments (the exact scores plotted in Figure~\ref{fig:bbh-small-large}).}
\label{tab:unfiltered-small-large}
\end{table*}

%% file: tables/unfiltered-short-long.tex
\begin{table*}[t]
\centering
\adjustbox{max width=\textwidth}{
\begin{tabular}{lcccccccc}
\toprule
& \multicolumn{4}{c}{\textbf{Short Teacher (32B)}} & \multicolumn{4}{c}{\textbf{Long Teacher (QwQ)}} \\
\cmidrule(lr){2-5} \cmidrule(lr){6-9}
\textbf{Task} & \textbf{0.5B} & \textbf{1.5B} & \textbf{3B} & \textbf{7B} & \textbf{0.5B} & \textbf{1.5B} & \textbf{3B} & \textbf{7B} \\
\midrule
Date Understanding & 53.0 & 77.0 & 80.0 & 93.0 & 67.0 & 84.0 & 88.0 & 89.0 \\
Dyck Languages & 12.0 & 35.0 & 56.0 & 56.0 & 5.0 & 18.0 & 26.0 & 48.0 \\
Formal Fallacies & 51.0 & 53.0 & 59.0 & 67.0 & 33.0 & 57.0 & 75.0 & 74.0 \\
Logical Deduction (3) & 34.0 & 81.0 & 95.0 & 95.0 & 50.0 & 85.0 & 91.0 & 97.0 \\
Logical Deduction (5) & 25.0 & 53.0 & 65.0 & 74.0 & 39.0 & 41.0 & 62.0 & 83.0 \\
Logical Deduction (7) & 14.0 & 38.0 & 39.0 & 60.0 & 22.0 & 44.0 & 46.0 & 57.0 \\
Penguins in a Table & 52.5 & 69.5 & 83.1 & 88.1 & 44.1 & 66.1 & 89.8 & 88.1 \\
Reasoning about Colored Objects & 33.0 & 71.0 & 82.0 & 91.0 & 46.0 & 77.0 & 85.0 & 96.0 \\
Ruin Names & 60.0 & 76.0 & 82.0 & 84.0 & 58.0 & 74.0 & 79.0 & 78.0 \\
Salient Translation Error Detection & 33.0 & 55.0 & 66.0 & 70.0 & 47.0 & 54.0 & 58.0 & 65.0 \\
Temporal Sequences & 75.0 & 90.0 & 99.0 & 100.0 & 67.0 & 88.0 & 98.0 & 99.0 \\
Tracking Shuffled Objects (3) & 95.0 & 100.0 & 95.0 & 99.0 & 94.0 & 100.0 & 100.0 & 98.0 \\
Tracking Shuffled Objects (5) & 87.0 & 93.0 & 100.0 & 100.0 & 85.0 & 99.0 & 98.0 & 99.0 \\
Tracking Shuffled Objects (7) & 68.0 & 90.0 & 100.0 & 99.0 & 57.0 & 99.0 & 99.0 & 99.0 \\
Word Sorting & 7.0 & 30.0 & 35.0 & 62.0 & 17.0 & 42.0 & 47.0 & 74.0 \\
\bottomrule
\end{tabular}
}
\caption{Per-task accuracy on BBH \textbf{without cross-teacher filtering} in the \emph{short--long} setting, i.e., our main experiments (the exact scores plotted in Figure~\ref{fig:bbh-short-long}).}
\label{tab:unfiltered-short-long}
\end{table*}

%% file: tables/filtering-small-large.tex
\begin{table*}[t]
\centering
\adjustbox{max width=\textwidth}{
\begin{tabular}{lcccccccc}
\toprule
& \multicolumn{4}{c}{\textbf{Small Teacher (14B)}} & \multicolumn{4}{c}{\textbf{Large Teacher (72B)}} \\
\cmidrule(lr){2-5} \cmidrule(lr){6-9}
\textbf{Task} & \textbf{0.5B} & \textbf{1.5B} & \textbf{3B} & \textbf{7B} & \textbf{0.5B} & \textbf{1.5B} & \textbf{3B} & \textbf{7B} \\
\midrule
Date Understanding & 51.0 & 81.0 & 91.0 & 92.0 & 55.0 & 85.0 & 80.0 & 91.0 \\
Dyck Languages & 3.0 & 12.0 & 28.0 & 37.0 & 1.0 & 12.0 & 23.0 & 43.0 \\
Formal Fallacies & 52.0 & 48.0 & 65.0 & 71.0 & 53.0 & 59.0 & 67.0 & 77.0 \\
Logical Deduction (3) & 45.0 & 76.0 & 92.0 & 95.0 & 39.0 & 83.0 & 93.0 & 96.0 \\
Logical Deduction (5) & 30.0 & 45.0 & 55.0 & 74.0 & 28.0 & 42.0 & 56.0 & 59.0 \\
Logical Deduction (7) & 17.0 & 37.0 & 49.0 & 60.0 & 13.0 & 40.0 & 41.0 & 58.0 \\
Penguins in a Table & 35.6 & 64.4 & 83.1 & 88.1 & 32.2 & 66.1 & 78.0 & 81.4 \\
Reasoning about Colored Objects & 35.0 & 74.0 & 85.0 & 90.0 & 25.0 & 76.0 & 84.0 & 87.0 \\
Ruin Names & 59.0 & 76.0 & 76.0 & 81.0 & 40.0 & 77.0 & 74.0 & 80.0 \\
Salient Translation Error Detection & 31.0 & 46.0 & 62.0 & 64.0 & 34.0 & 43.0 & 58.0 & 67.0 \\
Temporal Sequences & 82.0 & 96.0 & 97.0 & 98.0 & 65.0 & 96.0 & 99.0 & 100.0 \\
Tracking Shuffled Objects (3) & 85.0 & 99.0 & 98.0 & 99.0 & 83.0 & 99.0 & 99.0 & 100.0 \\
Tracking Shuffled Objects (5) & 85.0 & 94.0 & 97.0 & 100.0 & 77.0 & 94.0 & 99.0 & 99.0 \\
Tracking Shuffled Objects (7) & 41.0 & 94.0 & 99.0 & 98.0 & 43.0 & 93.0 & 100.0 & 99.0 \\
Word Sorting & 9.0 & 28.0 & 35.0 & 52.0 & 8.0 & 25.0 & 36.0 & 50.0 \\
\bottomrule
\end{tabular}
}
\caption{Per-task accuracy on BBH under \textbf{cross-teacher filtering} in the \emph{small--large} setting (see Table~\ref{tab:unfiltered-small-large} and Figure~\ref{fig:bbh-small-large} for unfiltered results).}
\label{tab:filtering-small-large}
\end{table*}

%% file: tables/filtering-short-long.tex
\begin{table*}[t]
\centering
\adjustbox{max width=\textwidth}{
\begin{tabular}{lcccccccc}
\toprule
& \multicolumn{4}{c}{\textbf{Short Teacher (32B)}} & \multicolumn{4}{c}{\textbf{Long Teacher (QwQ)}} \\
\cmidrule(lr){2-5} \cmidrule(lr){6-9}
\textbf{Task} & \textbf{0.5B} & \textbf{1.5B} & \textbf{3B} & \textbf{7B} & \textbf{0.5B} & \textbf{1.5B} & \textbf{3B} & \textbf{7B} \\
\midrule
Date Understanding & 52.0 & 80.0 & 83.0 & 92.0 & 55.0 & 75.0 & 85.0 & 84.0 \\
Dyck Languages & 1.0 & 7.0 & 24.0 & 38.0 & 1.0 & 10.0 & 31.0 & 25.0 \\
Formal Fallacies & 50.0 & 45.0 & 62.0 & 65.0 & 42.0 & 51.0 & 54.0 & 70.0 \\
Logical Deduction (3) & 48.0 & 73.0 & 88.0 & 96.0 & 47.0 & 82.0 & 93.0 & 96.0 \\
Logical Deduction (5) & 29.0 & 47.0 & 56.0 & 69.0 & 29.0 & 46.0 & 54.0 & 74.0 \\
Logical Deduction (7) & 10.0 & 41.0 & 38.0 & 62.0 & 18.0 & 36.0 & 47.0 & 67.0 \\
Penguins in a Table & 37.3 & 64.4 & 74.6 & 83.1 & 39.0 & 71.2 & 89.8 & 84.7 \\
Reasoning about Colored Objects & 31.0 & 72.0 & 85.0 & 90.0 & 41.0 & 77.0 & 86.0 & 94.0 \\
Ruin Names & 48.0 & 64.0 & 78.0 & 85.0 & 45.0 & 69.0 & 74.0 & 81.0 \\
Salient Translation Error Detection & 33.0 & 47.0 & 64.0 & 69.0 & 51.0 & 50.0 & 60.0 & 64.0 \\
Temporal Sequences & 81.0 & 81.0 & 95.0 & 97.0 & 71.0 & 72.0 & 100.0 & 100.0 \\
Tracking Shuffled Objects (3) & 95.0 & 100.0 & 100.0 & 100.0 & 94.0 & 99.0 & 99.0 & 98.0 \\
Tracking Shuffled Objects (5) & 82.0 & 98.0 & 99.0 & 99.0 & 89.0 & 99.0 & 99.0 & 99.0 \\
Tracking Shuffled Objects (7) & 63.0 & 99.0 & 99.0 & 100.0 & 64.0 & 94.0 & 97.0 & 99.0 \\
Word Sorting & 9.0 & 28.0 & 30.0 & 59.0 & 11.0 & 33.0 & 41.0 & 60.0 \\
\bottomrule
\end{tabular}
}
\caption{Per-task accuracy on BBH under \textbf{cross-teacher filtering} in the \emph{short--long} setting (see Table~\ref{tab:unfiltered-short-long} and Figure~\ref{fig:bbh-short-long} for unfiltered results).}
\label{tab:filtering-short-long}
\end{table*}

%% file: tables/bbh-pre-all.tex
\begin{table*}[htbp]
\centering
\adjustbox{max width=\textwidth}{
\begin{tabular}{lccccccccccc}
\toprule
& \multicolumn{5}{c}{\textbf{Student}} & \multicolumn{5}{c}{\textbf{Teacher}} & \\
\cmidrule(lr){2-6} \cmidrule(lr){7-11}
\textbf{Task} & \textbf{0.5B} & \textbf{1.5B} & \textbf{3B} & \textbf{7B} & \textbf{Avg.\ (S)} & \textbf{14B} & \textbf{32B} & \textbf{72B} & \textbf{QwQ} & \textbf{Avg.\ (T)} & \textbf{Gap (T$-$S)} \\
\midrule
Boolean Expressions & 75.2 & 88.0 & 88.8 & 96.4 & 87.1 & 99.2 & 100.0 & 100.0 & 96.8 & 99.0 & 11.9 \\
Causal Judgement & 48.7 & 59.9 & 54.5 & 60.4 & 55.9 & 65.8 & 69.0 & 71.7 & 67.9 & 68.6 & 12.7 \\
\rowcolor{backgray}
Date Understanding & 30.4 & 53.6 & 66.4 & 82.4 & 58.2 & 91.2 & 92.0 & 93.2 & 89.6 & 91.5 & 33.3 \\
Disambiguation QA & 31.2 & 49.2 & 63.6 & 69.2 & 53.3 & 82.4 & 79.2 & 82.8 & 74.4 & 79.7 & 26.4 \\
\rowcolor{backgray}
Dyck Languages & 0.0 & 0.0 & 3.6 & 11.2 & 3.7 & 31.6 & 58.8 & 61.2 & 40.4 & 48.0 & 44.3 \\
\rowcolor{backgray}
Formal Fallacies & 53.2 & 52.0 & 62.4 & 70.4 & 59.5 & 84.4 & 88.8 & 89.2 & 98.4 & 90.2 & 30.7 \\
Geometric Shapes & 22.0 & 28.8 & 49.2 & 66.4 & 41.6 & 54.4 & 58.4 & 54.8 & 70.8 & 59.6 & 18.0 \\
Hyperbaton & 55.2 & 72.8 & 74.4 & 90.0 & 73.1 & 99.6 & 98.8 & 99.2 & 99.6 & 99.3 & 26.2 \\
\rowcolor{backgray}
Logical Deduction (3) & 37.6 & 59.2 & 70.0 & 86.8 & 63.4 & 97.6 & 98.8 & 98.0 & 97.6 & 98.0 & 34.6 \\
\rowcolor{backgray}
Logical Deduction (5) & 22.4 & 33.2 & 45.6 & 59.2 & 40.1 & 80.4 & 88.4 & 92.0 & 90.0 & 87.7 & 47.6 \\
\rowcolor{backgray}
Logical Deduction (7) & 22.0 & 34.4 & 32.8 & 45.6 & 33.7 & 77.2 & 83.6 & 83.6 & 83.6 & 82.0 & 48.3 \\
Movie Recommendation & 37.2 & 48.4 & 74.0 & 66.4 & 56.5 & 74.0 & 78.0 & 80.4 & 87.6 & 80.0 & 23.5 \\
Multistep Arithmetic Two & 30.8 & 60.8 & 91.2 & 94.8 & 69.4 & 97.6 & 96.8 & 100.0 & 98.4 & 98.2 & 28.8 \\
Navigate & 58.4 & 73.6 & 90.0 & 95.2 & 79.3 & 98.0 & 98.8 & 99.2 & 97.2 & 98.3 & 19.0 \\
Object Counting & 53.6 & 58.8 & 63.2 & 78.4 & 63.5 & 92.4 & 90.0 & 97.6 & 89.6 & 92.4 & 28.9 \\
\rowcolor{backgray}
Penguins in a Table & 32.9 & 61.6 & 79.5 & 94.5 & 67.1 & 97.9 & 99.3 & 99.3 & 96.6 & 98.3 & 31.2 \\
\rowcolor{backgray}
Reasoning about Colored Objects & 20.8 & 62.0 & 75.6 & 90.0 & 62.1 & 95.2 & 98.4 & 99.2 & 99.6 & 98.1 & 36.0 \\
\rowcolor{backgray}
Ruin Names & 24.0 & 36.0 & 42.8 & 66.0 & 42.2 & 84.8 & 84.8 & 89.2 & 82.0 & 85.2 & 43.0 \\
\rowcolor{backgray}
Salient Translation Error Detection & 12.0 & 33.2 & 46.0 & 54.8 & 36.5 & 67.6 & 72.0 & 65.6 & 66.8 & 68.0 & 31.5 \\
Snarks & 46.6 & 57.3 & 64.6 & 79.8 & 62.1 & 84.3 & 87.1 & 91.6 & 93.3 & 89.0 & 27.0 \\
Sports Understanding & 48.8 & 73.2 & 82.0 & 87.2 & 72.8 & 90.4 & 90.4 & 94.8 & 91.2 & 91.7 & 18.9 \\
\rowcolor{backgray}
Temporal Sequences & 24.8 & 15.2 & 45.6 & 81.6 & 41.8 & 97.6 & 100.0 & 100.0 & 98.4 & 99.0 & 57.2 \\
\rowcolor{backgray}
Tracking Shuffled Objects (3) & 32.4 & 41.2 & 67.6 & 88.8 & 57.5 & 89.6 & 99.6 & 100.0 & 99.6 & 97.2 & 39.7 \\
\rowcolor{backgray}
Tracking Shuffled Objects (5) & 15.6 & 18.8 & 39.6 & 77.2 & 37.8 & 90.0 & 100.0 & 98.4 & 97.6 & 96.5 & 58.7 \\
\rowcolor{backgray}
Tracking Shuffled Objects (7) & 17.2 & 16.4 & 26.4 & 65.6 & 31.4 & 85.6 & 99.6 & 98.0 & 97.6 & 95.2 & 63.8 \\
Web of Lies & 59.6 & 80.8 & 100.0 & 99.2 & 84.9 & 99.6 & 100.0 & 100.0 & 100.0 & 99.9 & 15.0 \\
\rowcolor{backgray}
Word Sorting & 4.4 & 6.8 & 16.0 & 29.2 & 14.1 & 50.8 & 63.2 & 68.4 & 83.6 & 66.5 & 52.4 \\
\bottomrule
\end{tabular}
}
\caption{Results of few-shot ICL performance on BBH for \emph{pre-distillation} models (Qwen2.5-\{0.5B, 1.5B, 3B, 7B, 14B, 32B, 72B\}-Instruct and QwQ-32B-Preview). \textbf{Avg.} denotes the group average for student and teacher models, respectively. Tasks with a performance gap (\textbf{Gap} = Teacher Avg.\ $-$ Student Avg.) exceeding 30 percentage points are highlighted in gray; these tasks are selected for our distillation experiments.}
\label{tab:bbh-pre-all}
\end{table*}

%% file: tables/bbh-pre-all-gemma.tex
\begin{table*}[htbp]
\centering
\adjustbox{max width=\textwidth}{
\begin{tabular}{lccccccccc}
\toprule
& \multicolumn{5}{c}{\textbf{Student}} & \multicolumn{3}{c}{\textbf{Teacher (Gemma-2)}} & \\
\cmidrule(lr){2-6} \cmidrule(lr){7-9}
\textbf{Task} & \textbf{0.5B} & \textbf{1.5B} & \textbf{3B} & \textbf{7B} & \textbf{Avg.\ (S)} & \textbf{9B} & \textbf{27B} & \textbf{Avg.\ (T)} & \textbf{Gap (T$-$S)} \\
\midrule
Boolean Expressions & 75.2 & 88.0 & 88.8 & 96.4 & 87.1 & 91.6 & 91.2 & 91.4 & 4.3 \\
Causal Judgement & 48.7 & 59.9 & 54.5 & 60.4 & 55.9 & 63.1 & 66.8 & 65.0 & 9.1 \\
Date Understanding & 30.4 & 53.6 & 66.4 & 82.4 & 58.2 & 81.6 & 83.6 & 82.6 & 24.4 \\
Disambiguation QA & 31.2 & 49.2 & 63.6 & 69.2 & 53.3 & 75.6 & 82.4 & 79.0 & 25.7 \\
Dyck Languages & 0.0 & 0.0 & 3.6 & 11.2 & 3.7 & 9.2 & 9.2 & 9.2 & 5.5 \\
Formal Fallacies & 53.2 & 52.0 & 62.4 & 70.4 & 59.5 & 60.8 & 71.6 & 66.2 & 6.7 \\
Geometric Shapes & 22.0 & 28.8 & 49.2 & 66.4 & 41.6 & 52.0 & 54.8 & 53.4 & 11.8 \\
Hyperbaton & 55.2 & 72.8 & 74.4 & 90.0 & 73.1 & 77.2 & 98.4 & 87.8 & 14.7 \\
Logical Deduction (3) & 37.6 & 59.2 & 70.0 & 86.8 & 63.4 & 82.8 & 91.6 & 87.2 & 23.8 \\
Logical Deduction (5) & 22.4 & 33.2 & 45.6 & 59.2 & 40.1 & 57.2 & 61.2 & 59.2 & 19.1 \\
Logical Deduction (7) & 22.0 & 34.4 & 32.8 & 45.6 & 33.7 & 50.4 & 54.8 & 52.6 & 18.9 \\
Movie Recommendation & 37.2 & 48.4 & 74.0 & 66.4 & 56.5 & 59.6 & 73.2 & 66.4 & 9.9 \\
Multistep Arithmetic Two & 30.8 & 60.8 & 91.2 & 94.8 & 69.4 & 83.6 & 90.0 & 86.8 & 17.4 \\
Navigate & 58.4 & 73.6 & 90.0 & 95.2 & 79.3 & 58.8 & 98.0 & 78.4 & $-$0.9 \\
Object Counting & 53.6 & 58.8 & 63.2 & 78.4 & 63.5 & 88.0 & 95.6 & 91.8 & 28.3 \\
Penguins in a Table & 32.9 & 61.6 & 79.5 & 94.5 & 67.1 & 93.8 & 93.8 & 93.8 & 26.7 \\
Reasoning about Colored Objects & 20.8 & 62.0 & 75.6 & 90.0 & 62.1 & 79.6 & 88.4 & 84.0 & 21.9 \\
\rowcolor{backgray}
Ruin Names & 24.0 & 36.0 & 42.8 & 66.0 & 42.2 & 82.4 & 85.6 & 84.0 & 41.8 \\
Salient Translation Error Detection & 12.0 & 33.2 & 46.0 & 54.8 & 36.5 & 63.2 & 60.0 & 61.6 & 25.1 \\
Snarks & 46.6 & 57.3 & 64.6 & 79.8 & 62.1 & 80.9 & 87.1 & 84.0 & 21.9 \\
Sports Understanding & 48.8 & 73.2 & 82.0 & 87.2 & 72.8 & 92.4 & 94.4 & 93.4 & 20.6 \\
\rowcolor{backgray}
Temporal Sequences & 24.8 & 15.2 & 45.6 & 81.6 & 41.8 & 84.4 & 98.0 & 91.2 & 49.4 \\
Tracking Shuffled Objects (3) & 32.4 & 41.2 & 67.6 & 88.8 & 57.5 & 67.6 & 85.2 & 76.4 & 18.9 \\
\rowcolor{backgray}
Tracking Shuffled Objects (5) & 15.6 & 18.8 & 39.6 & 77.2 & 37.8 & 76.4 & 86.8 & 81.6 & 43.8 \\
\rowcolor{backgray}
Tracking Shuffled Objects (7) & 17.2 & 16.4 & 26.4 & 65.6 & 31.4 & 68.8 & 80.4 & 74.6 & 43.2 \\
Web of Lies & 59.6 & 80.8 & 100.0 & 99.2 & 84.9 & 100.0 & 100.0 & 100.0 & 15.1 \\
\rowcolor{backgray}
Word Sorting & 4.4 & 6.8 & 16.0 & 29.2 & 14.1 & 47.6 & 83.2 & 65.4 & 51.3 \\
\bottomrule
\end{tabular}
}
\caption{Results of few-shot ICL performance on BBH for \emph{pre-distillation} models (Qwen2.5-\{0.5B, 1.5B, 3B, 7B\}-Instruct as students and \textbf{Gemma-2-\{9B, 27B\}-it} as teachers). \textbf{Avg.} denotes the group average for student and teacher models, respectively. Tasks with a performance gap (\textbf{Gap} = Teacher Avg.\ $-$ Student Avg.) exceeding 30 percentage points are highlighted in gray; these tasks are selected for our distillation experiments.}
\label{tab:bbh-pre-all-gemma}
\end{table*}

%% file: custom.bib
@inproceedings{wei2022chain,
 author = {Wei, Jason and Wang, Xuezhi and Schuurmans, Dale and Bosma, Maarten and ichter, brian and Xia, Fei and Chi, Ed and Le, Quoc V and Zhou, Denny},
 booktitle = {Advances in Neural Information Processing Systems},
 editor = {S. Koyejo and S. Mohamed and A. Agarwal and D. Belgrave and K. Cho and A. Oh},
 pages = {24824--24837},
 publisher = {Curran Associates, Inc.},
 title = {Chain-of-Thought Prompting Elicits Reasoning in Large Language Models},
 url = {https://proceedings.neurips.cc/paper_files/paper/2022/file/9d5609613524ecf4f15af0f7b31abca4-Paper-Conference.pdf},
 volume = {35},
 year = {2022}
}

@inproceedings{kojima2022large,
 author = {Kojima, Takeshi and Gu, Shixiang (Shane) and Reid, Machel and Matsuo, Yutaka and Iwasawa, Yusuke},
 booktitle = {Advances in Neural Information Processing Systems},
 editor = {S. Koyejo and S. Mohamed and A. Agarwal and D. Belgrave and K. Cho and A. Oh},
 pages = {22199--22213},
 publisher = {Curran Associates, Inc.},
 title = {Large Language Models are Zero-Shot Reasoners},
 url = {https://proceedings.neurips.cc/paper_files/paper/2022/file/8bb0d291acd4acf06ef112099c16f326-Paper-Conference.pdf},
 volume = {35},
 year = {2022}
}

@inproceedings{li-etal-2023-symbolic,
    title = "Symbolic Chain-of-Thought Distillation: Small Models Can Also ``Think'' Step-by-Step",
    author = "Li, Liunian Harold  and
      Hessel, Jack  and
      Yu, Youngjae  and
      Ren, Xiang  and
      Chang, Kai-Wei  and
      Choi, Yejin",
    editor = "Rogers, Anna  and
      Boyd-Graber, Jordan  and
      Okazaki, Naoaki",
    booktitle = "Proceedings of the 61st Annual Meeting of the Association for Computational Linguistics (Volume 1: Long Papers)",
    month = jul,
    year = "2023",
    address = "Toronto, Canada",
    publisher = "Association for Computational Linguistics",
    url = "https://aclanthology.org/2023.acl-long.150/",
    doi = "10.18653/v1/2023.acl-long.150",
    pages = "2665--2679",
}

@InProceedings{pmlr-v202-fu23d,
  title = 	 {Specializing Smaller Language Models towards Multi-Step Reasoning},
  author =       {Fu, Yao and Peng, Hao and Ou, Litu and Sabharwal, Ashish and Khot, Tushar},
  booktitle = 	 {Proceedings of the 40th International Conference on Machine Learning},
  pages = 	 {10421--10430},
  year = 	 {2023},
  editor = 	 {Krause, Andreas and Brunskill, Emma and Cho, Kyunghyun and Engelhardt, Barbara and Sabato, Sivan and Scarlett, Jonathan},
  volume = 	 {202},
  series = 	 {Proceedings of Machine Learning Research},
  month = 	 {23--29 Jul},
  publisher =    {PMLR},
  url = 	 {https://proceedings.mlr.press/v202/fu23d.html},
}

@inproceedings{ho-etal-2023-large,
    title = "Large Language Models Are Reasoning Teachers",
    author = "Ho, Namgyu  and
      Schmid, Laura  and
      Yun, Se-Young",
    editor = "Rogers, Anna  and
      Boyd-Graber, Jordan  and
      Okazaki, Naoaki",
    booktitle = "Proceedings of the 61st Annual Meeting of the Association for Computational Linguistics (Volume 1: Long Papers)",
    month = jul,
    year = "2023",
    address = "Toronto, Canada",
    publisher = "Association for Computational Linguistics",
    url = "https://aclanthology.org/2023.acl-long.830/",
    doi = "10.18653/v1/2023.acl-long.830",
    pages = "14852--14882"
}

@inproceedings{magister-etal-2023-teaching,
    title = "Teaching Small Language Models to Reason",
    author = "Magister, Lucie Charlotte  and
      Mallinson, Jonathan  and
      Adamek, Jakub  and
      Malmi, Eric  and
      Severyn, Aliaksei",
    editor = "Rogers, Anna  and
      Boyd-Graber, Jordan  and
      Okazaki, Naoaki",
    booktitle = "Proceedings of the 61st Annual Meeting of the Association for Computational Linguistics (Volume 2: Short Papers)",
    month = jul,
    year = "2023",
    address = "Toronto, Canada",
    publisher = "Association for Computational Linguistics",
    url = "https://aclanthology.org/2023.acl-short.151/",
    doi = "10.18653/v1/2023.acl-short.151",
    pages = "1773--1781",
}

@inproceedings{hsieh-etal-2023-distilling,
    title = "Distilling Step-by-Step! Outperforming Larger Language Models with Less Training Data and Smaller Model Sizes",
    author = "Hsieh, Cheng-Yu  and
      Li, Chun-Liang  and
      Yeh, Chih-kuan  and
      Nakhost, Hootan  and
      Fujii, Yasuhisa  and
      Ratner, Alex  and
      Krishna, Ranjay  and
      Lee, Chen-Yu  and
      Pfister, Tomas",
    editor = "Rogers, Anna  and
      Boyd-Graber, Jordan  and
      Okazaki, Naoaki",
    booktitle = "Findings of the Association for Computational Linguistics: ACL 2023",
    month = jul,
    year = "2023",
    address = "Toronto, Canada",
    publisher = "Association for Computational Linguistics",
    url = "https://aclanthology.org/2023.findings-acl.507/",
    doi = "10.18653/v1/2023.findings-acl.507",
    pages = "8003--8017",
}

@inproceedings{shridhar-etal-2023-distilling,
    title = "Distilling Reasoning Capabilities into Smaller Language Models",
    author = "Shridhar, Kumar  and
      Stolfo, Alessandro  and
      Sachan, Mrinmaya",
    editor = "Rogers, Anna  and
      Boyd-Graber, Jordan  and
      Okazaki, Naoaki",
    booktitle = "Findings of the Association for Computational Linguistics: ACL 2023",
    month = jul,
    year = "2023",
    address = "Toronto, Canada",
    publisher = "Association for Computational Linguistics",
    url = "https://aclanthology.org/2023.findings-acl.441/",
    doi = "10.18653/v1/2023.findings-acl.441",
    pages = "7059--7073",
}

@inproceedings{hinton2015distilling,title	= {Distilling the Knowledge in a Neural Network},author	= {Geoffrey Hinton and Oriol Vinyals and Jeffrey Dean},year	= {2015},URL	= {http://arxiv.org/abs/1503.02531},booktitle	= {NIPS Deep Learning and Representation Learning Workshop}}

@inproceedings{kim-rush-2016-sequence,
    title = "Sequence-Level Knowledge Distillation",
    author = "Kim, Yoon  and
      Rush, Alexander M.",
    editor = "Su, Jian  and
      Duh, Kevin  and
      Carreras, Xavier",
    booktitle = "Proceedings of the 2016 Conference on Empirical Methods in Natural Language Processing",
    month = nov,
    year = "2016",
    address = "Austin, Texas",
    publisher = "Association for Computational Linguistics",
    url = "https://aclanthology.org/D16-1139/",
    doi = "10.18653/v1/D16-1139",
    pages = "1317--1327"
}

@misc{mukherjee2023orca,
      title={Orca: Progressive Learning from Complex Explanation Traces of GPT-4}, 
      author={Subhabrata Mukherjee and Arindam Mitra and Ganesh Jawahar and Sahaj Agarwal and Hamid Palangi and Ahmed Awadallah},
      year={2023},
      eprint={2306.02707v1},
      archivePrefix={arXiv},
      primaryClass={cs.CL},
      url={https://arxiv.org/abs/2306.02707}, 
}

@inproceedings{muennighoff-etal-2025-s1,
    title = "s1: Simple test-time scaling",
    author = "Muennighoff, Niklas  and
      Yang, Zitong  and
      Shi, Weijia  and
      Li, Xiang Lisa  and
      Fei-Fei, Li  and
      Hajishirzi, Hannaneh  and
      Zettlemoyer, Luke  and
      Liang, Percy  and
      Cand{\`e}s, Emmanuel  and
      Hashimoto, Tatsunori",
    editor = "Christodoulopoulos, Christos  and
      Chakraborty, Tanmoy  and
      Rose, Carolyn  and
      Peng, Violet",
    booktitle = "Proceedings of the 2025 Conference on Empirical Methods in Natural Language Processing",
    month = nov,
    year = "2025",
    address = "Suzhou, China",
    publisher = "Association for Computational Linguistics",
    url = "https://aclanthology.org/2025.emnlp-main.1025/",
    doi = "10.18653/v1/2025.emnlp-main.1025",
    pages = "20275--20321",
    ISBN = "979-8-89176-332-6",
}

@inproceedings{west-etal-2022-symbolic,
    title = "Symbolic Knowledge Distillation: from General Language Models to Commonsense Models",
    author = "West, Peter  and
      Bhagavatula, Chandra  and
      Hessel, Jack  and
      Hwang, Jena  and
      Jiang, Liwei  and
      Le Bras, Ronan  and
      Lu, Ximing  and
      Welleck, Sean  and
      Choi, Yejin",
    editor = "Carpuat, Marine  and
      de Marneffe, Marie-Catherine  and
      Meza Ruiz, Ivan Vladimir",
    booktitle = "Proceedings of the 2022 Conference of the North American Chapter of the Association for Computational Linguistics: Human Language Technologies",
    month = jul,
    year = "2022",
    address = "Seattle, United States",
    publisher = "Association for Computational Linguistics",
    url = "https://aclanthology.org/2022.naacl-main.341/",
    doi = "10.18653/v1/2022.naacl-main.341",
    pages = "4602--4625",
}

@inproceedings{li-etal-2025-small-models,
    title = "Small Models Struggle to Learn from Strong Reasoners",
    author = "Li, Yuetai  and
      Yue, Xiang  and
      Xu, Zhangchen  and
      Jiang, Fengqing  and
      Niu, Luyao  and
      Lin, Bill Yuchen  and
      Ramasubramanian, Bhaskar  and
      Poovendran, Radha",
    editor = "Che, Wanxiang  and
      Nabende, Joyce  and
      Shutova, Ekaterina  and
      Pilehvar, Mohammad Taher",
    booktitle = "Findings of the Association for Computational Linguistics: ACL 2025",
    month = jul,
    year = "2025",
    address = "Vienna, Austria",
    publisher = "Association for Computational Linguistics",
    url = "https://aclanthology.org/2025.findings-acl.1301/",
    doi = "10.18653/v1/2025.findings-acl.1301",
    pages = "25366--25394",
    ISBN = "979-8-89176-256-5",
}

@inproceedings{chen-etal-2025-unveiling-key,
    title = "Unveiling the Key Factors for Distilling Chain-of-Thought Reasoning",
    author = "Chen, Xinghao  and
      Sun, Zhijing  and
      Wenjin, Guo  and
      Zhang, Miaoran  and
      Chen, Yanjun  and
      Sun, Yirong  and
      Su, Hui  and
      Pan, Yijie  and
      Klakow, Dietrich  and
      Li, Wenjie  and
      Shen, Xiaoyu",
    editor = "Che, Wanxiang  and
      Nabende, Joyce  and
      Shutova, Ekaterina  and
      Pilehvar, Mohammad Taher",
    booktitle = "Findings of the Association for Computational Linguistics: ACL 2025",
    month = jul,
    year = "2025",
    address = "Vienna, Austria",
    publisher = "Association for Computational Linguistics",
    url = "https://aclanthology.org/2025.findings-acl.782/",
    doi = "10.18653/v1/2025.findings-acl.782",
    pages = "15094--15119",
    ISBN = "979-8-89176-256-5",
}

@misc{lopez-paz2016unifying,
      title={Unifying distillation and privileged information},
      author={David Lopez-Paz and L{\'e}on Bottou and Bernhard Sch{\"o}lkopf and Vladimir Vapnik},
      year={2016},
      eprint={1511.03643v3},
      archivePrefix={arXiv},
      primaryClass={stat.ML},
      url={https://arxiv.org/abs/1511.03643},
}

@inproceedings{jafari-etal-2021-annealing,
    title = "Annealing Knowledge Distillation",
    author = "Jafari, Aref  and
      Rezagholizadeh, Mehdi  and
      Sharma, Pranav  and
      Ghodsi, Ali",
    editor = "Merlo, Paola  and
      Tiedemann, Jorg  and
      Tsarfaty, Reut",
    booktitle = "Proceedings of the 16th Conference of the European Chapter of the Association for Computational Linguistics: Main Volume",
    month = apr,
    year = "2021",
    address = "Online",
    publisher = "Association for Computational Linguistics",
    url = "https://aclanthology.org/2021.eacl-main.212/",
    doi = "10.18653/v1/2021.eacl-main.212",
    pages = "2493--2504",
}

@article{mirzadeh2020improved, title={Improved Knowledge Distillation via Teacher Assistant}, volume={34}, url={https://ojs.aaai.org/index.php/AAAI/article/view/5963}, DOI={10.1609/aaai.v34i04.5963}, number={04}, journal={Proceedings of the AAAI Conference on Artificial Intelligence}, author={Mirzadeh, Seyed Iman and Farajtabar, Mehrdad and Li, Ang and Levine, Nir and Matsukawa, Akihiro and Ghasemzadeh, Hassan}, year={2020}, month={Apr.}, pages={5191-5198}}

@inproceedings{cho2019efficacy,
  title={On the efficacy of knowledge distillation},
  author={Cho, Jang Hyun and Hariharan, Bharath},
  booktitle={Proceedings of the IEEE/CVF international conference on computer vision},
  pages={4794--4802},
  year={2019},
  url={https://openaccess.thecvf.com/content_ICCV_2019/html/Cho_On_the_Efficacy_of_Knowledge_Distillation_ICCV_2019_paper.html}
}

@inproceedings{zhang-etal-2025-towards-law,
    title = "Towards the Law of Capacity Gap in Distilling Language Models",
    author = "Zhang, Chen  and
      Li, Qiuchi  and
      Song, Dawei  and
      Ye, Zheyu  and
      Gao, Yan  and
      Hu, Yao",
    editor = "Che, Wanxiang  and
      Nabende, Joyce  and
      Shutova, Ekaterina  and
      Pilehvar, Mohammad Taher",
    booktitle = "Proceedings of the 63rd Annual Meeting of the Association for Computational Linguistics (Volume 1: Long Papers)",
    month = jul,
    year = "2025",
    address = "Vienna, Austria",
    publisher = "Association for Computational Linguistics",
    url = "https://aclanthology.org/2025.acl-long.1097/",
    doi = "10.18653/v1/2025.acl-long.1097",
    pages = "22504--22528",
    ISBN = "979-8-89176-251-0",
}

@inproceedings{wang-etal-2023-scott,
    title = "{SCOTT}: Self-Consistent Chain-of-Thought Distillation",
    author = "Wang, Peifeng  and
      Wang, Zhengyang  and
      Li, Zheng  and
      Gao, Yifan  and
      Yin, Bing  and
      Ren, Xiang",
    editor = "Rogers, Anna  and
      Boyd-Graber, Jordan  and
      Okazaki, Naoaki",
    booktitle = "Proceedings of the 61st Annual Meeting of the Association for Computational Linguistics (Volume 1: Long Papers)",
    month = jul,
    year = "2023",
    address = "Toronto, Canada",
    publisher = "Association for Computational Linguistics",
    url = "https://aclanthology.org/2023.acl-long.304/",
    doi = "10.18653/v1/2023.acl-long.304",
    pages = "5546--5558",
}

@InProceedings{pmlr-v235-feng24e,
  title = 	 {Keypoint-based Progressive Chain-of-Thought Distillation for {LLM}s},
  author =       {Feng, Kaituo and Li, Changsheng and Zhang, Xiaolu and Zhou, Jun and Yuan, Ye and Wang, Guoren},
  booktitle = 	 {Proceedings of the 41st International Conference on Machine Learning},
  pages = 	 {13241--13255},
  year = 	 {2024},
  editor = 	 {Salakhutdinov, Ruslan and Kolter, Zico and Heller, Katherine and Weller, Adrian and Oliver, Nuria and Scarlett, Jonathan and Berkenkamp, Felix},
  volume = 	 {235},
  series = 	 {Proceedings of Machine Learning Research},
  month = 	 {21--27 Jul},
  publisher =    {PMLR},
  url = 	 {https://proceedings.mlr.press/v235/feng24e.html},
}

@article{liao2025neural, title={Neural-Symbolic Collaborative Distillation: Advancing Small Language Models for Complex Reasoning Tasks}, volume={39}, url={https://ojs.aaai.org/index.php/AAAI/article/view/34636}, DOI={10.1609/aaai.v39i23.34636}, number={23}, journal={Proceedings of the AAAI Conference on Artificial Intelligence}, author={Liao, Huanxuan and He, Shizhu and Xu, Yao and Zhang, Yuanzhe and Liu, Kang and Zhao, Jun}, year={2025}, month={Apr.}, pages={24567-24575}}

@inproceedings{chen-etal-2023-mcc,
    title = "{MCC}-{KD}: Multi-{C}o{T} Consistent Knowledge Distillation",
    author = "Chen, Hongzhan  and
      Wu, Siyue  and
      Quan, Xiaojun  and
      Wang, Rui  and
      Yan, Ming  and
      Zhang, Ji",
    editor = "Bouamor, Houda  and
      Pino, Juan  and
      Bali, Kalika",
    booktitle = "Findings of the Association for Computational Linguistics: EMNLP 2023",
    month = dec,
    year = "2023",
    address = "Singapore",
    publisher = "Association for Computational Linguistics",
    url = "https://aclanthology.org/2023.findings-emnlp.454/",
    doi = "10.18653/v1/2023.findings-emnlp.454",
    pages = "6805--6820",
}

@inproceedings{yan-etal-2025-towards,
    title = "Towards Efficient {C}o{T} Distillation: Self-Guided Rationale Selector for Better Performance with Fewer Rationales",
    author = "Yan, JianZhi  and
      Liu, Le  and
      Pan, Youcheng  and
      Chen, Shiwei  and
      Xiang, Yang  and
      Tang, Buzhou",
    editor = "Christodoulopoulos, Christos  and
      Chakraborty, Tanmoy  and
      Rose, Carolyn  and
      Peng, Violet",
    booktitle = "Findings of the Association for Computational Linguistics: EMNLP 2025",
    month = nov,
    year = "2025",
    address = "Suzhou, China",
    publisher = "Association for Computational Linguistics",
    url = "https://aclanthology.org/2025.findings-emnlp.413/",
    doi = "10.18653/v1/2025.findings-emnlp.413",
    pages = "7818--7835",
    ISBN = "979-8-89176-335-7",
}

@inproceedings{ye2025limo,
    title = {{LIMO}: Less is More for Reasoning},
    author = {Yixin Ye and Zhen Huang and Yang Xiao and Ethan Chern and Shijie Xia and Pengfei Liu},
    booktitle = {Second Conference on Language Modeling},
    year = {2025},
    url = {https://openreview.net/forum?id=T2TZ0RY4Zk},
}

@inproceedings{suzgun-etal-2023-challenging,
    title = "Challenging {BIG}-Bench Tasks and Whether Chain-of-Thought Can Solve Them",
    author = {Suzgun, Mirac  and
      Scales, Nathan  and
      Sch{\"a}rli, Nathanael  and
      Gehrmann, Sebastian  and
      Tay, Yi  and
      Chung, Hyung Won  and
      Chowdhery, Aakanksha  and
      Le, Quoc  and
      Chi, Ed  and
      Zhou, Denny  and
      Wei, Jason},
    editor = "Rogers, Anna  and
      Boyd-Graber, Jordan  and
      Okazaki, Naoaki",
    booktitle = "Findings of the Association for Computational Linguistics: ACL 2023",
    month = jul,
    year = "2023",
    address = "Toronto, Canada",
    publisher = "Association for Computational Linguistics",
    url = "https://aclanthology.org/2023.findings-acl.824/",
    doi = "10.18653/v1/2023.findings-acl.824",
    pages = "13003--13051",
}

@article{
srivastava2023beyond,
title={Beyond the Imitation Game: Quantifying and extrapolating the capabilities of language models},
author={Aarohi Srivastava and Abhinav Rastogi and Abhishek Rao and Abu Awal Md Shoeb and Abubakar Abid and Adam Fisch and Adam R. Brown and Adam Santoro and Aditya Gupta and Adri{\`a} Garriga-Alonso and Agnieszka Kluska and Aitor Lewkowycz and Akshat Agarwal and Alethea Power and Alex Ray and Alex Warstadt and Alexander W. Kocurek and Ali Safaya and Ali Tazarv and Alice Xiang and Alicia Parrish and Allen Nie and Aman Hussain and Amanda Askell and Amanda Dsouza and Ambrose Slone and Ameet Rahane and Anantharaman S. Iyer and Anders Johan Andreassen and Andrea Madotto and Andrea Santilli and Andreas Stuhlm{\"u}ller and Andrew M. Dai and Andrew La and Andrew Kyle Lampinen and Andy Zou and Angela Jiang and Angelica Chen and Anh Vuong and Animesh Gupta and Anna Gottardi and Antonio Norelli and Anu Venkatesh and Arash Gholamidavoodi and Arfa Tabassum and Arul Menezes and Arun Kirubarajan and Asher Mullokandov and Ashish Sabharwal and Austin Herrick and Avia Efrat and Aykut Erdem and Ayla Karaka{\c{s}} and B. Ryan Roberts and Bao Sheng Loe and Barret Zoph and Bart{\l}omiej Bojanowski and Batuhan {\"O}zyurt and Behnam Hedayatnia and Behnam Neyshabur and Benjamin Inden and Benno Stein and Berk Ekmekci and Bill Yuchen Lin and Blake Howald and Bryan Orinion and Cameron Diao and Cameron Dour and Catherine Stinson and Cedrick Argueta and Cesar Ferri and Chandan Singh and Charles Rathkopf and Chenlin Meng and Chitta Baral and Chiyu Wu and Chris Callison-Burch and Christopher Waites and Christian Voigt and Christopher D Manning and Christopher Potts and Cindy Ramirez and Clara E. Rivera and Clemencia Siro and Colin Raffel and Courtney Ashcraft and Cristina Garbacea and Damien Sileo and Dan Garrette and Dan Hendrycks and Dan Kilman and Dan Roth and C. Daniel Freeman and Daniel Khashabi and Daniel Levy and Daniel Mosegu{\'\i} Gonz{\'a}lez and Danielle Perszyk and Danny Hernandez and Danqi Chen and Daphne Ippolito and Dar Gilboa and David Dohan and David Drakard and David Jurgens and Debajyoti Datta and Deep Ganguli and Denis Emelin and Denis Kleyko and Deniz Yuret and Derek Chen and Derek Tam and Dieuwke Hupkes and Diganta Misra and Dilyar Buzan and Dimitri Coelho Mollo and Diyi Yang and Dong-Ho Lee and Dylan Schrader and Ekaterina Shutova and Ekin Dogus Cubuk and Elad Segal and Eleanor Hagerman and Elizabeth Barnes and Elizabeth Donoway and Ellie Pavlick and Emanuele Rodol{\`a} and Emma Lam and Eric Chu and Eric Tang and Erkut Erdem and Ernie Chang and Ethan A Chi and Ethan Dyer and Ethan Jerzak and Ethan Kim and Eunice Engefu Manyasi and Evgenii Zheltonozhskii and Fanyue Xia and Fatemeh Siar and Fernando Mart{\'\i}nez-Plumed and Francesca Happ{\'e} and Francois Chollet and Frieda Rong and Gaurav Mishra and Genta Indra Winata and Gerard de Melo and Germ{\`a}n Kruszewski and Giambattista Parascandolo and Giorgio Mariani and Gloria Xinyue Wang and Gonzalo Jaimovitch-Lopez and Gregor Betz and Guy Gur-Ari and Hana Galijasevic and Hannah Kim and Hannah Rashkin and Hannaneh Hajishirzi and Harsh Mehta and Hayden Bogar and Henry Francis Anthony Shevlin and Hinrich Schuetze and Hiromu Yakura and Hongming Zhang and Hugh Mee Wong and Ian Ng and Isaac Noble and Jaap Jumelet and Jack Geissinger and Jackson Kernion and Jacob Hilton and Jaehoon Lee and Jaime Fern{\'a}ndez Fisac and James B Simon and James Koppel and James Zheng and James Zou and Jan Kocon and Jana Thompson and Janelle Wingfield and Jared Kaplan and Jarema Radom and Jascha Sohl-Dickstein and Jason Phang and Jason Wei and Jason Yosinski and Jekaterina Novikova and Jelle Bosscher and Jennifer Marsh and Jeremy Kim and Jeroen Taal and Jesse Engel and Jesujoba Alabi and Jiacheng Xu and Jiaming Song and Jillian Tang and Joan Waweru and John Burden and John Miller and John U. Balis and Jonathan Batchelder and Jonathan Berant and J{\"o}rg Frohberg and Jos Rozen and Jose Hernandez-Orallo and Joseph Boudeman and Joseph Guerr and Joseph Jones and Joshua B. Tenenbaum and Joshua S. Rule and Joyce Chua and Kamil Kanclerz and Karen Livescu and Karl Krauth and Karthik Gopalakrishnan and Katerina Ignatyeva and Katja Markert and Kaustubh Dhole and Kevin Gimpel and Kevin Omondi and Kory Wallace Mathewson and Kristen Chiafullo and Ksenia Shkaruta and Kumar Shridhar and Kyle McDonell and Kyle Richardson and Laria Reynolds and Leo Gao and Li Zhang and Liam Dugan and Lianhui Qin and Lidia Contreras-Ochando and Louis-Philippe Morency and Luca Moschella and Lucas Lam and Lucy Noble and Ludwig Schmidt and Luheng He and Luis Oliveros-Col{\'o}n and Luke Metz and L{\"u}tfi Kerem Senel and Maarten Bosma and Maarten Sap and Maartje Ter Hoeve and Maheen Farooqi and Manaal Faruqui and Mantas Mazeika and Marco Baturan and Marco Marelli and Marco Maru and Maria Jose Ramirez-Quintana and Marie Tolkiehn and Mario Giulianelli and Martha Lewis and Martin Potthast and Matthew L Leavitt and Matthias Hagen and M{\'a}ty{\'a}s Schubert and Medina Orduna Baitemirova and Melody Arnaud and Melvin McElrath and Michael Andrew Yee and Michael Cohen and Michael Gu and Michael Ivanitskiy and Michael Starritt and Michael Strube and Micha{\l} Sw{\k{e}}drowski and Michele Bevilacqua and Michihiro Yasunaga and Mihir Kale and Mike Cain and Mimee Xu and Mirac Suzgun and Mitch Walker and Mo Tiwari and Mohit Bansal and Moin Aminnaseri and Mor Geva and Mozhdeh Gheini and Mukund Varma T and Nanyun Peng and Nathan Andrew Chi and Nayeon Lee and Neta Gur-Ari Krakover and Nicholas Cameron and Nicholas Roberts and Nick Doiron and Nicole Martinez and Nikita Nangia and Niklas Deckers and Niklas Muennighoff and Nitish Shirish Keskar and Niveditha S. Iyer and Noah Constant and Noah Fiedel and Nuan Wen and Oliver Zhang and Omar Agha and Omar Elbaghdadi and Omer Levy and Owain Evans and Pablo Antonio Moreno Casares and Parth Doshi and Pascale Fung and Paul Pu Liang and Paul Vicol and Pegah Alipoormolabashi and Peiyuan Liao and Percy Liang and Peter W Chang and Peter Eckersley and Phu Mon Htut and Pinyu Hwang and Piotr Mi{\l}kowski and Piyush Patil and Pouya Pezeshkpour and Priti Oli and Qiaozhu Mei and Qing Lyu and Qinlang Chen and Rabin Banjade and Rachel Etta Rudolph and Raefer Gabriel and Rahel Habacker and Ramon Risco and Rapha{\"e}l Milli{\`e}re and Rhythm Garg and Richard Barnes and Rif A. Saurous and Riku Arakawa and Robbe Raymaekers and Robert Frank and Rohan Sikand and Roman Novak and Roman Sitelew and Ronan Le Bras and Rosanne Liu and Rowan Jacobs and Rui Zhang and Russ Salakhutdinov and Ryan Andrew Chi and Seungjae Ryan Lee and Ryan Stovall and Ryan Teehan and Rylan Yang and Sahib Singh and Saif M. Mohammad and Sajant Anand and Sam Dillavou and Sam Shleifer and Sam Wiseman and Samuel Gruetter and Samuel R. Bowman and Samuel Stern Schoenholz and Sanghyun Han and Sanjeev Kwatra and Sarah A. Rous and Sarik Ghazarian and Sayan Ghosh and Sean Casey and Sebastian Bischoff and Sebastian Gehrmann and Sebastian Schuster and Sepideh Sadeghi and Shadi Hamdan and Sharon Zhou and Shashank Srivastava and Sherry Shi and Shikhar Singh and Shima Asaadi and Shixiang Shane Gu and Shubh Pachchigar and Shubham Toshniwal and Shyam Upadhyay and Shyamolima Shammie Debnath and Siamak Shakeri and Simon Thormeyer and Simone Melzi and Siva Reddy and Sneha Priscilla Makini and Soo-Hwan Lee and Spencer Torene and Sriharsha Hatwar and Stanislas Dehaene and Stefan Divic and Stefano Ermon and Stella Biderman and Stephanie Lin and Stephen Prasad and Steven Piantadosi and Stuart Shieber and Summer Misherghi and Svetlana Kiritchenko and Swaroop Mishra and Tal Linzen and Tal Schuster and Tao Li and Tao Yu and Tariq Ali and Tatsunori Hashimoto and Te-Lin Wu and Th{\'e}o Desbordes and Theodore Rothschild and Thomas Phan and Tianle Wang and Tiberius Nkinyili and Timo Schick and Timofei Kornev and Titus Tunduny and Tobias Gerstenberg and Trenton Chang and Trishala Neeraj and Tushar Khot and Tyler Shultz and Uri Shaham and Vedant Misra and Vera Demberg and Victoria Nyamai and Vikas Raunak and Vinay Venkatesh Ramasesh and vinay uday prabhu and Vishakh Padmakumar and Vivek Srikumar and William Fedus and William Saunders and William Zhang and Wout Vossen and Xiang Ren and Xiaoyu Tong and Xinran Zhao and Xinyi Wu and Xudong Shen and Yadollah Yaghoobzadeh and Yair Lakretz and Yangqiu Song and Yasaman Bahri and Yejin Choi and Yichi Yang and Sophie Hao and Yifu Chen and Yonatan Belinkov and Yu Hou and Yufang Hou and Yuntao Bai and Zachary Seid and Zhuoye Zhao and Zijian Wang and Zijie J. Wang and Zirui Wang and Ziyi Wu},
journal={Transactions on Machine Learning Research},
issn={2835-8856},
year={2023},
url={https://openreview.net/forum?id=uyTL5Bvosj},
note={Featured Certification}
}

@inproceedings{agarwal2024manyshot,
 author = {Agarwal, Rishabh and Singh, Avi and Zhang, Lei and Bohnet, Bernd and Rosias, Luis and Chan, Stephanie and Zhang, Biao and Anand, Ankesh and Abbas, Zaheer and Nova, Azade and Co-Reyes, John D. and Chu, Eric and Behbahani, Feryal and Faust, Aleksandra and Larochelle, Hugo},
 booktitle = {Advances in Neural Information Processing Systems},
 editor = {A. Globerson and L. Mackey and D. Belgrave and A. Fan and U. Paquet and J. Tomczak and C. Zhang},
 pages = {76930--76966},
 publisher = {Curran Associates, Inc.},
 title = {Many-Shot In-Context Learning},
 url = {https://proceedings.neurips.cc/paper_files/paper/2024/file/8cb564df771e9eacbfe9d72bd46a24a9-Paper-Conference.pdf},
 volume = {37},
 year = {2024}
}

@inproceedings{honda-etal-2025-distilling,
    title = "Distilling Many-Shot In-Context Learning into a Cheat Sheet",
    author = "Honda, Ukyo  and
      Murakami, Soichiro  and
      Zhang, Peinan",
    editor = "Christodoulopoulos, Christos  and
      Chakraborty, Tanmoy  and
      Rose, Carolyn  and
      Peng, Violet",
    booktitle = "Findings of the Association for Computational Linguistics: EMNLP 2025",
    month = nov,
    year = "2025",
    address = "Suzhou, China",
    publisher = "Association for Computational Linguistics",
    url = "https://aclanthology.org/2025.findings-emnlp.930/",
    doi = "10.18653/v1/2025.findings-emnlp.930",
    pages = "17158--17178",
    ISBN = "979-8-89176-335-7",
}

@inproceedings{hendrycks2021measuring,
 author = {Hendrycks, Dan and Burns, Collin and Kadavath, Saurav and Arora, Akul and Basart, Steven and Tang, Eric and Song, Dawn and Steinhardt, Jacob},
 booktitle = {Proceedings of the Neural Information Processing Systems Track on Datasets and Benchmarks},
 editor = {J. Vanschoren and S. Yeung},
 pages = {},
 title = {Measuring Mathematical Problem Solving With the MATH Dataset},
 url = {https://datasets-benchmarks-proceedings.neurips.cc/paper_files/paper/2021/file/be83ab3ecd0db773eb2dc1b0a17836a1-Paper-round2.pdf},
 volume = {1},
 year = {2021}
}

@misc{cobbe2021training,
      title={Training Verifiers to Solve Math Word Problems}, 
      author={Karl Cobbe and Vineet Kosaraju and Mohammad Bavarian and Mark Chen and Heewoo Jun and Lukasz Kaiser and Matthias Plappert and Jerry Tworek and Jacob Hilton and Reiichiro Nakano and Christopher Hesse and John Schulman},
      year={2021},
      eprint={2110.14168v2},
      archivePrefix={arXiv},
      primaryClass={cs.LG},
      url={https://arxiv.org/abs/2110.14168}, 
}

@inproceedings{he-etal-2024-olympiadbench,
    title = "{O}lympiad{B}ench: A Challenging Benchmark for Promoting {AGI} with Olympiad-Level Bilingual Multimodal Scientific Problems",
    author = "He, Chaoqun  and
      Luo, Renjie  and
      Bai, Yuzhuo  and
      Hu, Shengding  and
      Thai, Zhen  and
      Shen, Junhao  and
      Hu, Jinyi  and
      Han, Xu  and
      Huang, Yujie  and
      Zhang, Yuxiang  and
      Liu, Jie  and
      Qi, Lei  and
      Liu, Zhiyuan  and
      Sun, Maosong",
    editor = "Ku, Lun-Wei  and
      Martins, Andre  and
      Srikumar, Vivek",
    booktitle = "Proceedings of the 62nd Annual Meeting of the Association for Computational Linguistics (Volume 1: Long Papers)",
    month = aug,
    year = "2024",
    address = "Bangkok, Thailand",
    publisher = "Association for Computational Linguistics",
    url = "https://aclanthology.org/2024.acl-long.211/",
    doi = "10.18653/v1/2024.acl-long.211",
    pages = "3828--3850",
}

@misc{qwen2025qwen25,
      title={Qwen2.5 Technical Report}, 
      author={Qwen and : and An Yang and Baosong Yang and Beichen Zhang and Binyuan Hui and Bo Zheng and Bowen Yu and Chengyuan Li and Dayiheng Liu and Fei Huang and Haoran Wei and Huan Lin and Jian Yang and Jianhong Tu and Jianwei Zhang and Jianxin Yang and Jiaxi Yang and Jingren Zhou and Junyang Lin and Kai Dang and Keming Lu and Keqin Bao and Kexin Yang and Le Yu and Mei Li and Mingfeng Xue and Pei Zhang and Qin Zhu and Rui Men and Runji Lin and Tianhao Li and Tianyi Tang and Tingyu Xia and Xingzhang Ren and Xuancheng Ren and Yang Fan and Yang Su and Yichang Zhang and Yu Wan and Yuqiong Liu and Zeyu Cui and Zhenru Zhang and Zihan Qiu},
      year={2025},
      eprint={2412.15115v2},
      archivePrefix={arXiv},
      primaryClass={cs.CL},
      url={https://arxiv.org/abs/2412.15115}, 
}

@inproceedings{zheng-etal-2024-llamafactory,
    title = "{L}lama{F}actory: Unified Efficient Fine-Tuning of 100+ Language Models",
    author = "Zheng, Yaowei  and
      Zhang, Richong  and
      Zhang, Junhao  and
      Ye, Yanhan  and
      Luo, Zheyan",
    editor = "Cao, Yixin  and
      Feng, Yang  and
      Xiong, Deyi",
    booktitle = "Proceedings of the 62nd Annual Meeting of the Association for Computational Linguistics (Volume 3: System Demonstrations)",
    month = aug,
    year = "2024",
    address = "Bangkok, Thailand",
    publisher = "Association for Computational Linguistics",
    url = "https://aclanthology.org/2024.acl-demos.38/",
    doi = "10.18653/v1/2024.acl-demos.38",
    pages = "400--410",
}

@inproceedings{
hu2022lora,
title={Lo{RA}: Low-Rank Adaptation of Large Language Models},
author={Edward J Hu and yelong shen and Phillip Wallis and Zeyuan Allen-Zhu and Yuanzhi Li and Shean Wang and Lu Wang and Weizhu Chen},
booktitle={International Conference on Learning Representations},
year={2022},
url={https://openreview.net/forum?id=nZeVKeeFYf9}
}

@misc{eval-harness,
  author       = {Gao, Leo and Tow, Jonathan and Abbasi, Baber and Biderman, Stella and Black, Sid and DiPofi, Anthony and Foster, Charles and Golding, Laurence and Hsu, Jeffrey and Le Noac'h, Alain and Li, Haonan and McDonell, Kyle and Muennighoff, Niklas and Ociepa, Chris and Phang, Jason and Reynolds, Laria and Schoelkopf, Hailey and Skowron, Aviya and Sutawika, Lintang and Tang, Eric and Thite, Anish and Wang, Ben and Wang, Kevin and Zou, Andy},
  title        = {The Language Model Evaluation Harness},
  month        = 07,
  year         = 2024,
  publisher    = {Zenodo},
  version      = {v0.4.3},
  doi          = {10.5281/zenodo.12608602},
  url          = {https://zenodo.org/records/12608602}
}

@misc{gemmateam2024gemma2,
      title={Gemma 2: Improving Open Language Models at a Practical Size}, 
      author={Gemma Team and Morgane Riviere and Shreya Pathak and Pier Giuseppe Sessa and Cassidy Hardin and Surya Bhupatiraju and Léonard Hussenot and Thomas Mesnard and Bobak Shahriari and Alexandre Ramé and Johan Ferret and Peter Liu and Pouya Tafti and Abe Friesen and Michelle Casbon and Sabela Ramos and Ravin Kumar and Charline Le Lan and Sammy Jerome and Anton Tsitsulin and Nino Vieillard and Piotr Stanczyk and Sertan Girgin and Nikola Momchev and Matt Hoffman and Shantanu Thakoor and Jean-Bastien Grill and Behnam Neyshabur and Olivier Bachem and Alanna Walton and Aliaksei Severyn and Alicia Parrish and Aliya Ahmad and Allen Hutchison and Alvin Abdagic and Amanda Carl and Amy Shen and Andy Brock and Andy Coenen and Anthony Laforge and Antonia Paterson and Ben Bastian and Bilal Piot and Bo Wu and Brandon Royal and Charlie Chen and Chintu Kumar and Chris Perry and Chris Welty and Christopher A. Choquette-Choo and Danila Sinopalnikov and David Weinberger and Dimple Vijaykumar and Dominika Rogozińska and Dustin Herbison and Elisa Bandy and Emma Wang and Eric Noland and Erica Moreira and Evan Senter and Evgenii Eltyshev and Francesco Visin and Gabriel Rasskin and Gary Wei and Glenn Cameron and Gus Martins and Hadi Hashemi and Hanna Klimczak-Plucińska and Harleen Batra and Harsh Dhand and Ivan Nardini and Jacinda Mein and Jack Zhou and James Svensson and Jeff Stanway and Jetha Chan and Jin Peng Zhou and Joana Carrasqueira and Joana Iljazi and Jocelyn Becker and Joe Fernandez and Joost van Amersfoort and Josh Gordon and Josh Lipschultz and Josh Newlan and Ju-yeong Ji and Kareem Mohamed and Kartikeya Badola and Kat Black and Katie Millican and Keelin McDonell and Kelvin Nguyen and Kiranbir Sodhia and Kish Greene and Lars Lowe Sjoesund and Lauren Usui and Laurent Sifre and Lena Heuermann and Leticia Lago and Lilly McNealus and Livio Baldini Soares and Logan Kilpatrick and Lucas Dixon and Luciano Martins and Machel Reid and Manvinder Singh and Mark Iverson and Martin Görner and Mat Velloso and Mateo Wirth and Matt Davidow and Matt Miller and Matthew Rahtz and Matthew Watson and Meg Risdal and Mehran Kazemi and Michael Moynihan and Ming Zhang and Minsuk Kahng and Minwoo Park and Mofi Rahman and Mohit Khatwani and Natalie Dao and Nenshad Bardoliwalla and Nesh Devanathan and Neta Dumai and Nilay Chauhan and Oscar Wahltinez and Pankil Botarda and Parker Barnes and Paul Barham and Paul Michel and Pengchong Jin and Petko Georgiev and Phil Culliton and Pradeep Kuppala and Ramona Comanescu and Ramona Merhej and Reena Jana and Reza Ardeshir Rokni and Rishabh Agarwal and Ryan Mullins and Samaneh Saadat and Sara Mc Carthy and Sarah Cogan and Sarah Perrin and Sébastien M. R. Arnold and Sebastian Krause and Shengyang Dai and Shruti Garg and Shruti Sheth and Sue Ronstrom and Susan Chan and Timothy Jordan and Ting Yu and Tom Eccles and Tom Hennigan and Tomas Kocisky and Tulsee Doshi and Vihan Jain and Vikas Yadav and Vilobh Meshram and Vishal Dharmadhikari and Warren Barkley and Wei Wei and Wenming Ye and Woohyun Han and Woosuk Kwon and Xiang Xu and Zhe Shen and Zhitao Gong and Zichuan Wei and Victor Cotruta and Phoebe Kirk and Anand Rao and Minh Giang and Ludovic Peran and Tris Warkentin and Eli Collins and Joelle Barral and Zoubin Ghahramani and Raia Hadsell and D. Sculley and Jeanine Banks and Anca Dragan and Slav Petrov and Oriol Vinyals and Jeff Dean and Demis Hassabis and Koray Kavukcuoglu and Clement Farabet and Elena Buchatskaya and Sebastian Borgeaud and Noah Fiedel and Armand Joulin and Kathleen Kenealy and Robert Dadashi and Alek Andreev},
      year={2024},
      eprint={2408.00118v3},
      archivePrefix={arXiv},
      primaryClass={cs.CL},
      url={https://arxiv.org/abs/2408.00118}, 
}
